\documentclass{article} 
\usepackage{tcolorbox} 
\usepackage{amsmath,amsfonts}
\usepackage{myICLR,times}

\usepackage{bm}
\usepackage{xfrac}
\usepackage{amssymb}
\usepackage{array}
\usepackage{booktabs}
\usepackage{multirow}
\usepackage{comment}
\usepackage{subcaption}
\usepackage{authblk}
\usepackage{xcolor} 
\usepackage{graphicx} 
\usepackage{geometry} 
\usepackage{enumitem}
\usepackage{url}
\newcommand{\zr}[1]{\textcolor{cyan}{[ZR: #1]}}

\usepackage{hyperref}
\usepackage{tikz}
\usetikzlibrary{positioning,fit,arrows.meta}
\usetikzlibrary{backgrounds}

\def\eqref#1{equation~\ref{#1}}

\def\1{\bm{1}}

\DeclareMathAlphabet{\mathsfit}{\encodingdefault}{\sfdefault}{m}{sl}
\SetMathAlphabet{\mathsfit}{bold}{\encodingdefault}{\sfdefault}{bx}{n}

\def\gD{{\mathcal{D}}}

\def\gL{{\mathcal{L}}}

\def\gN{{\mathcal{N}}}
\def\gO{{\mathcal{O}}}

\def\sR{{\mathbb{R}}}

\newcommand{\E}{\mathbb{E}}

\newcommand{\R}{\mathbb{R}}

\DeclareMathOperator*{\ext}{ext}

\DeclareMathOperator{\Tr}{Tr}

\title{Mitigating the Curse of Detail: Scaling Arguments for Feature Learning and Sample Complexity}
\author[1]{Noa Rubin}
\author[2]{Orit Davidovich}
\author[1]{Zohar Ringel}
\affil[1]{Racah Institute of Physics\\
         Hebrew University\\
         Jerusalem, Israel}
\affil[2]{IBM Research\\ Haifa, Israel.}
\usepackage{fancyhdr}
\iclrfinalcopy 
\begin{document}

\maketitle

\begin{abstract}
Two pressing topics in the theory of deep learning are the interpretation of feature learning (FL) mechanisms and the determination of implicit bias of networks in the rich regime. Current theories of rich FL often appear in the form of high-dimensional non-linear equations, which require computationally intensive numerical solutions. Given the many details that go into defining a deep learning problem, this analytical complexity is a significant and often unavoidable challenge. Here, we propose a powerful heuristic route for predicting the data and width scales at which various patterns of FL emerge. This form of scale analysis is considerably simpler than such exact theories and reproduces the scaling exponents of various known results. In addition, we make novel predictions on complex toy architectures, such as three-layer non-linear networks and attention heads, thus extending the scope of first-principle theories of deep learning.
\end{abstract}

\section{Introduction}

There is a clear need for a better theoretical understanding of deep learning. However, efforts to construct such theories inevitably suffer from a ``curse of details''. Indeed, since any choice of architecture, activation, data measure, and training protocol affects performance, finding a theory with true predictive power that accurately accounts for all those details is unlikely. One workaround is to focus on analytically tractable toy models, an approach that can often uncover interesting fundamental aspects. However, analytical tractability is a fragile, fine-tuned property; thus, a large explainability gap remains between such toy models and more complex data/architecture settings. 

An alternative approach focuses on scaling properties of neural networks, which appear more robust. Two well-established examples are empirically predicting network performance by extrapolating learning curves using power laws \cite{kaplan2020scaling,hestness2017deeplearningscalingpredictable}, and providing theory-inspired suggestions for hyperparameter transfer techniques \cite{yang2022tensorprogramsvtuning,bordelon2023depthwise}. Indeed, it is often the case \cite{cardy1996scaling} that predicting scaling exponents is easier than predicting exact or approximate behaviors. As a simple toy model of this, consider the integral $\int_{-\infty}^{\infty} dx g(x/P)$. While $g(\cdot)$ needs to be fine-tuned for exact computations, a change of variable reveals a robust linear scaling with $P$ for any $g(\cdot)$. 


This work focuses on scaling properties of feature learning (FL). This phenomenon, and, more generally, interpretability, has been studied extensively both from the practical and theoretical sides. On the practical side, mechanistic interpretability \cite{bereska2024mechanisticinterpretabilityaisafety} has provided us with statistical explanations for why some predictions are made and the underlying decision mechanisms. On the theory side, kernel approaches \cite{bordelon2022self, aitchison2021deep} Saad and Solla type approaches \cite{Saad1995,Bruno2023,bietti2022learning} (and their Bayesian counterparts \cite{cui2025highdimensionallearningnarrowneural}) allow us to solve simple non-linear teacher-student networks in the rich regime. 

This work examines feature learning through a Bayesian lens, where it is often defined as a deviation from the Gaussian Process (GP) prior. This deviation has been shown to manifest in several distinct regime dependent ways, including: (i) updates to the kernel while the distribution remains Gaussian \cite{LiSompolinsky2021,rubin_kernels_2025,ariosto2022statistical,seroussi2023separation, aitchison2019bigger}; (ii) the distribution changes from zero mean-Gaussian to a mixture of Gaussians \cite{rubin_grokking_2024}; and (iii) the specialization of specific neurons \cite{barbier_statistical_2025}; as well as combinations of the above \cite{ringel_applications_2025, meegen_coding_2024}. However, the frameworks capturing these FL effects are often mathematically cumbersome and restricted to shallow networks. Even when extended to deep architectures \cite{barbier_statistical_2025, li2020statistical}, their computational complexity prevents the derivation of intuitive scaling laws. Ultimately, this difficulty in computation compounds with the context-specificity of these mechanisms, complicating unified comparison of scaling laws across varied architectural settings.   


In this work, we introduce a novel heuristic framework addressing the challenging task of making first-principles predictions on sample complexity and feature learning effects in Bayesian NNs. We consider Bayesian NNs as these are a standard proxy for SGD-trained networks at equilibrium \cite{mingard2020sgd,mandt2017stochastic}. The heuristic nature of our approach (see Fig. \ref{Fig:LogiFlow} for a schematic overview), allows us to bypass otherwise highly complex analysis and extract insights using simple pen-and-paper calculations, resulting in non-trivial predictions across a broad range of architectures and settings. We indeed validate our framework for two- and three-layer fully connected networks (using ReLU and Erf activations), attention heads, and convolutional neural networks, covering both teacher-student, regression, and classification tasks. Across these diverse settings, our main results are:
\begin{enumerate}[topsep=0pt, itemsep=0pt, parsep=0pt, leftmargin=*]
    \item \textbf{Predicting feature learning patterns:} We predict which of the FL phenomena from the literature discussed above will emerge in different architectures and across varying scales of architectural variables, such as input dimension, layer width, regularization, and parametrization choices such as mean-field versus standard scaling. We map transitions between these patterns and derive the explicit scaling laws governing their strength.
    \item \textbf{Predicting sample complexity scale:} We determine the scaling behavior of $P_{*}$, the threshold sample size at which learning is possible, as a function of the architectural variables and parametrization choices.
\end{enumerate}

\begin{figure}[t]
\centering
\newlength{\GroupH}
\setlength{\GroupH}{27mm} 

\resizebox{\linewidth}{!}{%
\begin{tikzpicture}[>=Stealth, node distance=8mm, every node/.style={font=\footnotesize}]
\tikzset{
  box/.style={draw=black, rounded corners, fill=gray!30, very thick,
              minimum width=22mm, minimum height=8mm, align=center},
  group/.style={draw=black, rounded corners, fill=gray!10, dotted, line width=1pt, inner sep=4mm}
}

\node[box]                      (b1) {Learning certificate:\\ $A_f$ (bounds MSE)};
\node[box, right=5.5mm of b1]     (b2) {Upper bound \\on posterior \\ $\pi[A_f\geq\alpha]$};
\node[box, right=5.5mm of b2]     (b3) {Prior $A_f$ rate\\  function: $E(\alpha)$  } ;
\node[box, right=6mm of b3]     (b4) {Lower bound \\on sample size \\ $P\gtrsim E(\alpha)$};
\node[box, right=9mm of b4]    (b5) {Variational\\ $E(\alpha)\approx\tilde{E}_{q_*}(\alpha)$};
\node[box, right=9mm of b5]    (b6) {Feature \\Learning\\Patterns};
\node[box, right=5.5mm of b6]     (b7) {Kernel \\Feature\\Propagation};

\draw[->, very thick] (b1) -- node[above]{}                                        node[below]{} (b2);
\draw[->, very thick] (b2) -- node[above]{} 
                              node[below]{} (b3);
\draw[->, very thick] (b3) -- node[above, font=\fontsize{7}{7}\selectfont]{LDT} 
                              node[below]{} (b4);
\draw[->, very thick] (b4) -- node[above]{} 
                              node[below]{} (b5);
\draw[->, very thick] (b5) -- node[above]{} 
                              node[below]{} (b6);

\draw[->, very thick] (b6) -- node[above]{} 
                              node[below]{} (b7);

\begin{pgfonlayer}{background}
  \node[group, fit=(b1)(b2)(b3)(b4), inner ysep=0pt, inner xsep=3.5mm, minimum height=\GroupH] (g1) {};
  \node[group, fit=(b5),          inner ysep=0pt, inner xsep=3.5mm, minimum height=\GroupH] (g2) {};
  \node[group, fit=(b6)(b7),      inner ysep=0pt, inner xsep=3.5mm, minimum height=\GroupH] (g3) {};
\end{pgfonlayer}

\node[anchor=north west, font=\small, fill=none, inner sep=5pt, draw=none]
  at (g1.north west) {\textbf{Bounds}};
\node[anchor=north west, font=\small, fill=none, inner sep=5pt]
  at (g2.north west) {\textbf{Approximations}};
\node[anchor=north west, font=\small, fill=none, inner sep=5pt]
  at (g3.north west) {\textbf{Heuristics}};
\end{tikzpicture}%
}
\caption{Logical flow of sample complexity derivation.
{\bf Bounds:} {\bf ({\romannumeral 1})} we lower bound the test MSE by $(1-A_f)^2$, where $A_f$ is the alignment of output and target; {\bf ({\romannumeral 2})} we establish an upper bound on the probability to observe good learning (i.e., strong alignment $A_f \geq  \alpha \approx 1$)  in the posterior using the negative-log-probability of the rare event of good learning in the prior; {\bf ({\romannumeral 3})} we define the energy $E(\alpha)$ as the negative log-Chernoff bound on the \textit{prior} probability of successful learning (rate function); {\bf ({\romannumeral 4})} we leverage $E(\alpha)$ to bound the minimal sample size necessary for good learning in the \textit{posterior}; 
\textbf{Approximations:} \textbf{(\romannumeral 5)} Since the exact bound $E(\alpha)$ is intractable, we derive a variational approximation using kernel-adaptation techniques to provide an explicit formula (Sec.~\ref{sec:variational_analysis}).
\textbf{Heuristics:} \textbf{(\romannumeral 6)}--\textbf{(\romannumeral 7)} we propose ``feature learning patterns'' as heuristic variational candidates, selecting the pattern $q_*$ that minimizes the approximation and utilizing heuristic scaling relations to model how feature amplification propagates through downstream layers (Sec.~\ref{sec:heuristics}).
}

\label{Fig:LogiFlow}
\end{figure}


\section{Setup}
   
We consider here several types of feedforward networks, but, for the sake of clarity, we illustrate the main derivation on deep fully connected networks (FCNs) and, later, when we analyze specific problems, augment it for convolutional neural networks (CNNs) as needed. Our FCNs are defined by 
\begin{align}
f(x)=\sum_{i=1}^{N_{L-1}}w^L_i\sigma(h_i^{L-1}(x)),\quad
 \text{where}\ \  h_i^{l>1}(x)=\sum_{j=1}^{N_{l-1}}W_{ij}^{l}\sigma(h_j^{l-1}(x)),\ \  
h_j^1(x)=\sum_{k=1}^d W_{jk}^{1}x_k,
\label{eq:pre-activations}
\end{align}
where $\sigma$ can be any activation function, and we refer to $h^{l}_i$'s as {\it pre-acitvations}. 
We consider Bayesian neural networks, as Bayesian descriptions are a commonly used proxy for network behavior after long-time stochastic training \cite{wilson_bayesian_2020,wilson_case_2020,naveh2021predicting}. Alternatively, they represent an exact solution to Langevin dynamics with weight decay \cite{welling_bayesian_nodate}. We denote the target function by $y$ and the training sample size by $P$, and assume training with Mean Squared Error (MSE) loss. The quadratic weight decay for each layer $l$ is set to $\kappa N_{l-1}/\sigma_l^2$, where $\kappa$ is the ridge parameter and $N_0=d$ is the input dimension. This choice of weight decay results in a Gaussian prior distribution for the weights given by $W_{i,j}^{l}\sim\mathcal{N}(0,\sigma_l^2/N_{l-1})$ for $i=1,..,N_{l}$ , $j=1,...,N_{l-1}$. The possible outputs $f$ of such a network, given $y$ and $P$, are then distributed according to the posterior:
    \begin{equation}
        \pi\left(f \mid y,P \right)=\frac{1}{Z}\exp\left(-\frac{1}{2\kappa}\sum_{\nu=1}^P \left[f(x_{\nu})-y(x_{\nu})\right]^{2}\right)p_0(f),
    \end{equation}
where $Z$ is the normalization constant, $\{x_{\nu}\}_{\nu=1}^{P}$ is the training dataset of size $P$, and $p_0(f)$ is the prior defined as $p_0(f)=\int d\Theta p\left(\Theta\right)\delta\left[f-f_{\Theta}\right]$, determined by the weight decay. Here, $\Theta$ is the collection of all network weights, and $p\left(\Theta\right)$ corresponds to the density of the prior weight distribution, which we take to be  Gaussian with a diagonal covariance, representing quadratic weight decay, and $f_{\Theta}$ is the network architecture with weights $\Theta$. We further set $p(\Theta)$ such that pre-activations are all $\gO(1)$ under the prior \cite{He2015}. For classification tasks, see App.~\ref{app:classification}. As a measure for learning, we consider an observable, which we refer to as {\it alignment}, given by 
\begin{equation}
A_{f}:=\langle f,y\rangle / \langle y,y\rangle,
\end{equation}
where $\langle g,h \rangle = \int d\mu_x g(x)h(x)$ is the functional inner product, and $d \mu_x$ is some test measure, which, conveniently, does not need to be the measure from which the training set was drawn. We similarly define $\langle g,K,h\rangle=\int d\mu_x d\mu_{x'} g(x)K(x,x')h(x')$ for any kernel $K$. Alignment represents the extent to which the network learns a function that is proportional to the target. It bounds the test MSE via the Cauchy–Schwarz inequality $\int d\mu_x (f(x)-y(x))^2 \geq \langle y,y \rangle (A_f-1)^2$. Having $A_f \approx 1$ is thus a necessary condition for successful learning.

\section{Alignment and Sample Complexity}
\label{sec:bounds_alignment}
We turn to analyze sample complexity via an upper bound on the probability of finding $A_f\geq \alpha$ for $\alpha \approx 1$. We begin with a  theoretical bound on the posterior that mainly depends on the chance that a random network, chosen from the prior, produces an alignment of at least $\alpha$. We denote the prior and posterior alignment probabilities by $\Pr_{p_{0}}\left[A_{f}\geq\alpha\right]$ and $\Pr_{\pi}\left[A_{f}\geq\alpha\right]$ respectively.
Following simple arguments (see App.~\ref{app:upper_bounds} and App.~\ref{app:classification} for generalization to classification.), we obtain the following bound on the log posterior \footnote{See also \cite{lavie_demystifying_2025}, for a similar data-agnostic bound in the context of lazy learning.}
\begin{equation}
\log\left(\Pr\nolimits_{\pi}\left[A_{f}\geq\alpha\right]\right) < Pk/\left(2\kappa\right)+\log\left(\Pr\nolimits_{p_0}\left[A_{f}\geq\alpha\right]\right),
\end{equation}
where $k = P^{-1} \sum_{\nu=1}^P \mathbb{E}_{p_0}[ (f(x_{\nu})-y(x_\nu))^2]$ is the only training-set dependent quantity and is generally of order one. The Bayesian interpretation of successful learning is having $\Pr\nolimits_{\pi}\left[A_{f}\geq\alpha\right] \approx \gO(1)$. 
\footnote{More concretely, we can assume some tolerance $\epsilon$ and then $\pi\left(A_f \approx 1  \right)=(\epsilon)^{-1} \int_{0}^{\epsilon} d\epsilon \pi(A_f=1-\epsilon)$}. 
Since a random network is unlikely to achieve strong alignment, $\log\Pr_{p_{0}}\left[A_{f}\geq\alpha\right]$ would typically be highly negative for large $\alpha$. Therefore, a sufficiently large data term is required to cancel this effect. Explicitly,  
\begin{align}
P \gtrsim -2\kappa\log\Pr\nolimits_{p_{0}}\left[A_{f}\geq\alpha\right]/k.
\label{eq:P_star_with_P_alpha}
\end{align}
Thus, up to the ridge parameter and the $\mathcal{O}(1)$ factor $k$, depending on the training set, the log probability of prior alignment with the target lower bounds the sample complexity. Here, it is worth noting that the bound becomes tight when overfitting effects are small, which is typically the case for $ k/\kappa \sim \mathcal{O}(1)$. Taking $\kappa \rightarrow 0$ encourages overfitting (though often benignly \cite{bartlett_benign_2020}) and trivializes this bound. We conjecture that, in this case, $\kappa$ should be kept $\mathcal{O}(1)$ based on the effective ridge treatment \cite{canatar_spectral_2021,cohen_learning_2021,bartlett_benign_2020}. Establishing this conjecture is outside the scope of this work. From a PAC-Bayesian perspective, an analogous bound would require $P$ to be much larger than the KL-divergence between the prior and posterior\footnote{Following the data-processing-inequality, one can lower-bound the KL-divergence between the full prior and posterior probabilities by the KL-divergence of a coarser probability of an $A_f \geq \alpha$ event in the prior and posterior. The latter KL divergence is given by $-\log\Pr\nolimits_{p_{0}}\left[A_{f}\geq\alpha\right]$} (e.g., \cite{10.1145/307400.307435}). 
More recently, prior-posterior relations have been studied in the context of complex Boolean functions \cite{mingard2025characterisinginductivebiasesneural}. 

Following the Chernoff inequality, we can find an upper bound for the probability (and a lower bound for $P$) via
\begin{equation}
  P\geq - 2\kappa/k\ \log\Pr\nolimits_{p_{0}}\left[A_{f}\geq\alpha\right] \geq 2\kappa/k\ E(\alpha), \ \ \ E(\alpha)=-\log \inf_{t>0}e^{-t\alpha}\mathbb{E}_{p_0}\left[e^{tA_{f}}\right].
\end{equation}
Where we refer to $E(\alpha)$ as the \textit{energy}. We can thus express the minimal sample size necessary for learning, $P_*$, through the energy as $P_*\propto  E(\alpha)$. In App. \ref{app:upper_bound_chern_FCN}, we provide an asymptotically exact solution for $E(\alpha)$, and compute it explicitly for a two layer network. We also argue and demonstrate that our bound is inherently tied to FL. Indeed, a network sampled from the prior that achieves such alignment is a statistical outlier, driven by the emergence of an internal structure which mimics FL (see also Fig. \ref{fig:2_layer_exp}). Nevertheless, such a direct LDT approach is computationally prohibitive in most cases of interest. We therefore introduce a heuristic LDT-based method for evaluating $P_*$. This method not only enables predicting the scaling of $P_*$ but also the FL effects that lead to successful learning.

In this section, we adopt a variational approach to estimating $P_*$ by comparing different modes of feature learning 
\footnote{Viewed here formally as emergent weight/pre-activation structures enabling the outlier.} 
under a certain loss (see (\ref{eq:est_E}) below). While many approaches predict different FL mechanisms \cite{pacelli_statistical_2023,fischer_critical_2024,meegen_coding_2024,buzaglo_how_2025,LiSompolinsky2021,seroussi_separation_2023,rubin_kernels_2025,rubin_grokking_2024,barbier_statistical_2025}, they are often case-dependent, highly detailed, and complex. Thus, we propose a method that abstracts key FL mechanisms from these frameworks into distinct, 
comparable patterns. 

\section{Variational Analysis}\label{sec:variational_analysis}
Our next objective is to make the sample complexity bound tractable. This requires estimating the prior probability term, $\Pr_{p_{0}}\left[A_{f}\geq\alpha\right]$, for alignments $\alpha \approx 1$. As a first step, we simplify this by relating the cumulative distribution function to the probability density denoted by $p_{A_f}(\alpha)$. As shown in App.~\ref{app:upper_bound_chern}, for large alignments, we have $E(\alpha)\approx-\log p_{A_f}(\alpha)$. This allows us to re-express $P_*$ in terms of the density: $P_*= -2\kappa \log p_{A_f}(\alpha)/k$. However, computing $p_{A_f}(\alpha)$ directly remains intractable. We therefore turn to a variational approach to estimate it. As explained in the next section, we wish to express the variational probability density in terms of pre-activations $h$ (\ref{eq:pre-activations}). Accordingly, in App. \ref{app:layerwise_seperation}, we follow standard statistical mechanics techniques to express this density as
\vspace*{-0.7em}
\begin{align}
     p_{A_{f}}(\alpha) = \int \mathcal{D}h \, \mathcal{N}\left(\alpha \mid 0, \langle y, \tilde{K}_{L-1}, y \rangle \right) \prod_{l=1}^{L-1} \prod_{i=1}^{N_{l}} \mathcal{N}\left(h_{i}^{l} \mid 0, \tilde{K}_{l-1}\right)
\end{align}
Where for each $l$, the kernels $\tilde{K}_{l-1}$ themselves depend on the preactivations of layer $l-1$, and similarly the fluctuations in $\alpha$ depend on the preactivations of the penultimate layer, given by
\vspace*{-0.80em}
\begin{equation}\label{eq:kernel_def}
\tilde{K}_{l>0}(x,x') = \frac{\sigma_{l+1}^{2}}{N_{l}} \sum_{i=1}^{N_{l}} \sigma\left(h_{i}^{l}(x)\right)\sigma\left(h_{i}^{l}(x')\right), \quad K_{0}(x,x') = \frac{\sigma_{1}^{2}}{d} x \cdot x' 
\end{equation}

Using statistical mechanics based notation, we define the Hamiltonian $H_{p,\alpha}$ and fluctuating "partition function" $Z_{A_{f}}$
\vspace*{-1.5em}
\begin{align}
    \label{eq:H_def_main_text}
    H_{p,\alpha}\left(h\right)&=\frac{\alpha^{2}}{2\left\langle y,\tilde{K}_{L-1},y\right\rangle }+\frac{1}{2}\sum_{l=1}^{L-1}\sum_{i=1}^{N_{l}}\left\langle h_{i}^{l},\tilde{K}_{l-1}^{-1},h_{i}^{l}\right\rangle ,\\\log Z_{A_{f}}\left(h\right)&=\log\left\langle y,\tilde{K}_{L-1},y\right\rangle +\sum_{l=1}^{L-1}\Tr\log\tilde{K}_{l-1}.\nonumber
\end{align}
Consequently, $p_{A_{f}}(\alpha)$ takes the form
 $p_{A_{f}}(\alpha) = \int \mathcal{D}h \, \exp\left(-H_{p,\alpha}(h) - \log Z_{A_f}(h)\right),$ where the integrand defines an effective measure on the preactivation space for a given $\alpha$. We next approximate this measure per $\alpha$ by an analytically tractable variational estimate, $q$. We follow a similar convention here: for any $q, \alpha$, we define $q_{\alpha}(h) := Z_{q,\alpha}^{-1} e^{-H_{q,\alpha}(h)}$, requiring that the minimum value of $H_{q,\alpha}(h)$ be zero. For a Gaussian $q$, this reduces exactly to Eq.~\ref{eq:H_def_main_text}, with kernels $\tilde{K}_{l-1}$ that are independent of $h$. The variational computation follows by looking for $q_{\alpha}(h)$ that minimizes the KL divergence between the measure on $h$ which defines $p_{A_f}$, and $q$. The KL divergence can also be used in the estimation of $E(\alpha):=-\log(p_{A_f}(\alpha))$, following the  Feynman–Bogoliubov inequality \cite{kuzemsky_variational_2015,bogolubov_introduction_2009,clark_variational_1968}. Here we provide a brief description -- for the full derivation see App.~\ref{app:variation_est_intro}.
By applying the Feynman–Bogoliubov inequality, we obtain an upper bound on $E(\alpha)$ 
\begin{align}
E(\alpha)\approx \min_{q_{\alpha}}(\mathbb{E}_{h\sim q_{\alpha}}[\log(Z_{A_f}(h)/Z_{q,\alpha})]+\tilde{E}_q(\alpha)),\quad \quad \tilde{E}_q(\alpha) = \mathbb{E}_{h\sim q_{\alpha}}[H_{p,\alpha}(h)-H_{q,\alpha} (h)] \label{eq:variational_energy}
\end{align}
We argue in App. \ref{app:justification_for} that for $\alpha\approx 1$, the log terms are subleading w.r.t. $\tilde{E}_q(\alpha)$. Defining $q_{*,\alpha}$ to be the measure that minimizes $\tilde{E}_{q}(\alpha)$, we obtain $E(\alpha)\approx \tilde{E}_{q_*}(\alpha)$. 
Next, we turn to estimating the variational energy $\tilde{E}_q(\alpha)$ Eq.~(\ref{eq:variational_energy}) for $\alpha \sim 1$, omitting all $\alpha$ indices for brevity. In App. \ref{app:layerwise_seperation} we simplify $p_{A_f}$, and show that the distribution in each layer depends only on the previous through a fluctuating non-linear operator. Next, we assume that this kernel is weakly fluctuating, and replace it with its expectation w.r.t. the variational distribution. This choice of approximation aligns with various works on deep non-linear networks, where layer-wise kernels are identified as the relevant and sufficient set of order parameters \cite{rubin_kernels_2025,fischer_critical_2024,seroussi_separation_2023,ringel_applications_2025}. We further take a decoupled Gaussian variational ansatz so that $q(h)=\prod_{l=1}^{L-1}\prod_{i=1}^{N_l}q_{l,i}(h_i^l)$ where $q_{l,i}$ is Gaussian with mean $\mu_{l,i}$ and variance $Q_{l,i}$. As shown in App.~\ref{app:layerwise_var}, the variational energy estimate is then given by 
\begin{equation}\label{eq:est_E}
\tilde{E}_q 
\propto \sum_{l=1}^{L-1}\sum_{i=1}^{N_{l}}\underbrace{\left(\mathbb{E}_{h\sim\mathcal{N}(\mu_{l,i},Q_{l,i})}\left[\left\langle h,K_{l-1}^{-1}-Q_{l,i}^{-1},h\right\rangle \right]+\left\langle \mu_{l,i},Q_{l,i}^{-1},\mu_{l,i}\right\rangle \right)}_{=:\Delta_{l,i}}
+ \underbrace{\langle y, K_{L-1}, y \rangle^{-1}}_{=:a_y},
\end{equation}
where we define $K_l=\mathbb{E}_{h\sim q}[\tilde{K}_l]$. Here, the $\Delta_{l,i}$ terms arise from the difference between the approximated kernel and the actual one, and the $a_y$ term results from enforcing an alignment $\alpha\approx 1$. Requiring that $q$ minimize $\tilde{E}_q$ and $\alpha\approx 1$, we estimate $\tilde{E}_q\propto E(\alpha\approx 1)\propto P_*$. Another interpretation of $\Delta_{l,i}$, discussed in App. In \ref{App:WeightSpace}, the excess weight is due to FL. This viewpoint is useful for FL patterns involving circuits, as the latter have a sharp imprint in weight-space. The above kernel viewpoint is, however, more general and can be used both for circuits and for more distributed learning patterns.  



\section{Heuristics for Manual Computation of Variational approximation}
\label{sec:heuristics}
\subsection{Feature Learning Patterns}\label{sec:feature_learning}
While the above variational approach allows a variety of candidate $q$'s, we focus on the previously mentioned set of feature-learning scenarios that have been extensively studied in the literature. Although this subset may appear restrictive, by varying behaviors among layers and between different neurons of the same layer, it already captures a wide range of phenomena. 
We then need to compare the variational energy ($\tilde{E}_q$), as detailed in Sec.~\ref{sec:variational_analysis}, for such combinations and select the minimizer. The optimal pattern is an indication of the FL that emerges in the network to enable strong alignment, as motivated in App.~\ref{app:upper_bound_chern}.  Concretely, per layer and neuron pre-activation ($h^l_{i}(x)$), we allow one of the following choices:

{\bf (1) Gaussian Process (GP).}
Here, $q_{l,i}$ is a Gaussian process (GP) so that $h_{l,i}\sim \mathcal{N}(0,K_{l-1})$ with $K_{l-1}$ defined as the expectation of the kernel defined in (\ref{eq:kernel_def}).
This choice defines the ``base model'' of FL. For FCNs \footnote{For CNNs, even in the lazy regime, deeper kernels have a different input scope and hence generate a new structure.}, it implies that the network propagates feature structure forward without altering latent features (see Sec.~\ref{sec:kernel_prop}). When all layers and neurons follow this distribution, the network reduces to the neural network GP (NNGP) \cite{neal_bayesian_1996}, where no FL occurs. Introducing any of the patterns below in a subset of neurons enables FL to emerge.

{\bf (2) Gaussian Feature Learning (GFL).}
In this scenario, pre-activations remain Gaussian with zero mean, but the covariance is modified relative to the GP scenario (1): the kernel of the previous layer is amplified by a factor $D$ in the direction of a specific feature (e.g.,  an eigenfunction of $K_{l-1}$) $\Phi_*^l$, one may also consider generalizations to several features). Thus, here too, the distribution is a GP but with a different covariance $Q_{l,i}$ given by
\begin{equation}
Q_{l,i}(x,x')=K_{l-1}(x,x')+D\langle \Phi_*^l,K_{l-1},\Phi_*^l\rangle\,\Phi_*^l(x)\Phi_*^l(x').
\end{equation}

{\bf (3) Specialization.}
In this scenario, a given neuron specializes to a particular feature $\Phi_*^l$ with proportionality constant $\mu_{l,i}$. This pattern corresponds to a Gaussian distribution which is sharply peaked around a non-zero mean $\mu_{l,i}\Phi_*^l$ \footnote{Taking equilibrated networks and increasing the amount of data, specialization was shown in \cite{rubin_grokking_2024}  to emerge as a first-order phase transition where the average of preactivations suddenly shifts to $\mu_{l,i}$. This behavior was further associated with Grokking, suggesting a potential specialization-grokking link. }. Explicitly, we define the distribution of the specialized neuron as 
\begin{equation}
    q_{l,i}(\langle h_i^l,\Phi_*^l\rangle)=\delta[\langle h_i^l,\Phi_{*}^l\rangle-\mu_{l,i}],\quad  q_{l,i}(\langle h_i^l,\Phi_{\perp}^l\rangle)=\mathcal{N}(0,\langle\phi_{\perp}^l ,K_{l-1},\Phi_{\perp}^l\rangle).
\end{equation}

\subsection{Layer-wise Feature Propagation}
\label{sec:kernel_prop}

Since the variational energy of each layer depends on the kernel of the previous layer, an important element in our heuristic is understanding how the choice of pattern in a given layer affects the kernel and its spectrum in the subsequent layer. To this end, we define feature learning as any deviation from the baseline GP pattern (see Sec.~\ref{sec:feature_learning}), such as introducing a non-zero mean to the distribution (i.e., specialization) or altering its covariance structure (i.e., GFL). In our framework, a "feature" refers either to the mean $\mu_{l,i}$ of $q_{l,i}$ or to an eigenfunction of its covariance operator $Q_{l,i}(x,x')$. 

We now outline several key claims concerning how features typically propagate between layers in FCNs. In this context, we consider a data measure that is i.i.d. Gaussian with zero mean and variance $1$, not because it approximates the data well, but rather because it provides an unbiased baseline (see also \cite{lavie_demystifying_2025}) for measuring function overlaps. Depending on the input, other choices can also be considered (e.g.,  permutation-symmetric measures over discrete tokens \cite{lavie_towards_2024}). The following claims with their justifications should be understood as heuristic principles or rationalizations of empirically observed phenomena. Proving them in general or augmenting for different architectures is left for future work. For further details and empirical results, see App.~\ref{app:feature_prop}.

\textbf{Claim ({\romannumeral 1}): Neuron specialization creates a spectral spike.} Assume that $M$ neurons in layer $l$ specialize on a single feature $\Phi_*^l(x)$, the subsequent kernel $K_{l}$ develops a new, dominant spectral feature corresponding to $\sigma(\Phi^l_*(x))$. The corresponding RKHS norm of this feature is amplified, scaling as $\mathcal{O}(N_{l}/M)$. \textbf{Justification:} When $M$ neurons specialize, the next layer's kernel is approximately $K_l(x,x') = A(x,x') + \tfrac{M}{N_l}\sigma(\Phi^l_{*}(x))\sigma(\Phi^l_{*}(x'))$, where $A$ is the contribution from the non-specialized neurons. Treating the specialization term as a rank-1 update, the Sherman-Morrison formula shows that its RKHS norm becomes $(R_A^{-1} + M /N)^{-1}$, where $R_A$  is the RKHS norm of $A$, which satisfies $R_A^{-1}\ll M /N$ in typical high-dimensional settings. 

\textbf{Claim ({\romannumeral 2}): Amplified features in the pre-activation kernel create amplified higher-order features in the post-activation kernel.}  If a feature $\Phi^l_*(x)$ in kernel $K_{l}$ has its eigenvalue enhanced by a factor $D$ (i.e., $\lambda_*\rightarrow \lambda_*D$), then the corresponding $m$-th order power of this feature  $(\Phi^l_{*})^m(x)$ will have the bulk of its spectral decomposition, under the downstream kernel, shifted up by $D^m$, with similar effect on the inverse RKHS norm. 
\textbf{Justification:} A Taylor expansion of $K_{l+1}$ in terms of the eigenfunctions of $K_{l}$ shows that the term corresponding to $(\Phi^l_{*})^m(x)$ will have a coefficient scaling with $(\lambda_{*}D)^m$. We argue that this term is difficult to span using other terms in this expansion, allowing us to treat it as a spectral spike and analyze it similarly to Claim (i). A numerical demonstration of this effect is shown in Fig.~\ref{fig:FeaturePropagation_GFL}. 

\textbf{Claim ({\romannumeral 3}): Lazy layers preserve the relative scale of features from the previous layer.}
In the absence of FL, a properly normalized lazy layer approximately preserves the eigenspectrum of the previous kernel. If a feature  $\Phi^l_*(x)$ has an eigenvalue $\lambda_*$ with respect to the pre-activation kernel given by $K_{l-1}$, its effective eigenvalue with respect to the post-activation kernel $K_l$ will also be proportional to $\lambda_*$. \textbf{Justification:} Follows from Claim ({\romannumeral 2}) taking $D=1$.

\begin{tcolorbox}[title=Propagation rules for FCNs]
  (1) \textbf{Specialization}: Layer $l$ specialized $M$ neurons on $\Phi_*^l$. For any feature $\Phi$ satisfying $\langle \sigma(\Phi_*^l), \Phi \rangle \neq 0$, we approximate $ \langle \Phi, K_{l}^{-1}, \Phi \rangle \propto \left[ \sum_{i \text{ sp.}} \frac{\mu_{i,l}^2}{N_l} \right]^{-1}$, where we sum over all specializing neurons.

(2) \textbf{GFL}: Layer $l$ amplified fluctuation along  $\Phi_*^l$ by $D$ so that $ \langle \Phi_*^l, K_{l}^{-1}, \Phi_*^l \rangle=(D\lambda_*)^{-1}$, where $\lambda_*$ is the GP value of the inner product.  Then for any $m$ we have 
$\langle (\Phi_*^l)^m, K_{l}^{-1}, (\Phi_*^l)^m \rangle \propto (D\lambda_*)^{-m}$.
 \end{tcolorbox}
 
\section{Concrete Examples}

\begin{figure}
    \centering
    \begin{tikzpicture}

    \node[anchor=south west, inner sep=0] (image) at (0,0) {\includegraphics[width=1\linewidth]{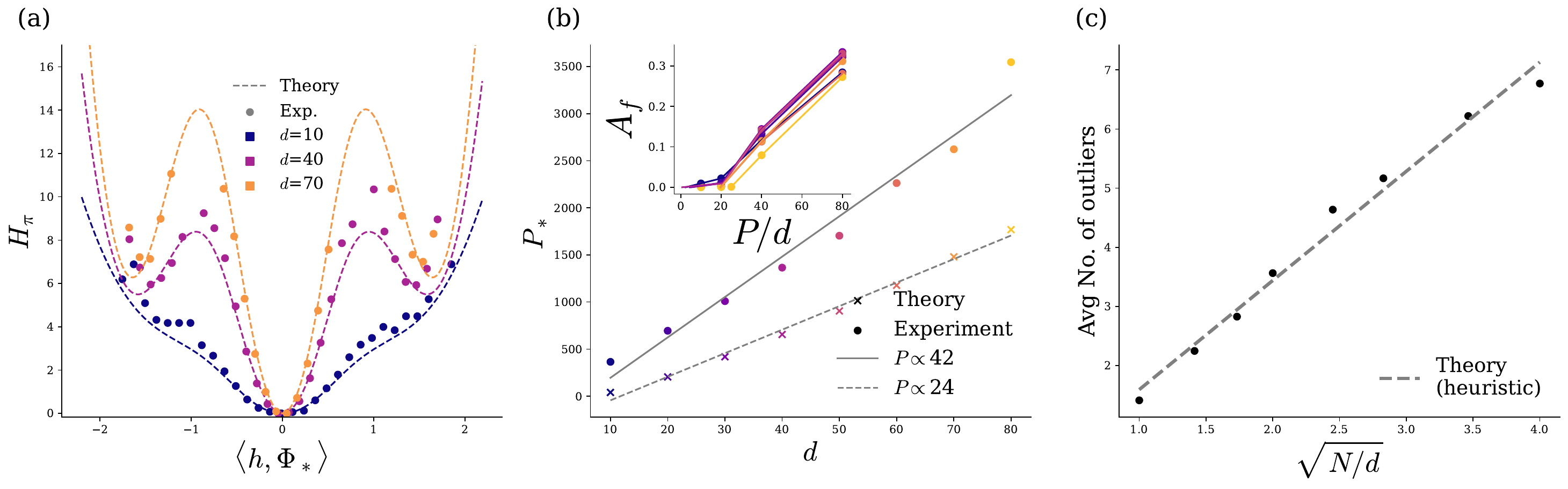}};

    \begin{scope}[x={(image.south east)}, y={(image.north west)}]
    \node[anchor=center, font=\fontsize{6}{12}\selectfont] at (0.615, 0.21) {$d$};
    \node[anchor=center, font=\fontsize{6}{12}\selectfont] at (0.615, 0.27) {$d$};
    \end{scope}
    \end{tikzpicture}
    \caption{Numerical and experimental results for a two-layer erf network trained on the normalized third Hermite polynomial ($m=3$). In panel (a) we compare the experimental results and exact theoretical predictions (computed utilizing LDT, see App. \ref{app:upper_bound_chern_FCN}) for the distribution of the alignment of the hidden layer pre-activation with the linear feature. Here we follow the same notation as in (~\ref{eq:H_def_main_text}), so that $H_{\pi}$ is the negative log posterior of the preactivations up to an additive constant that enforces zero minimum. We also find the pre-activation distribution corresponds to $q(h)$ for $q\sim\text{M-Sp}$, as predicted by our heuristic approach. Panel (b) compares theoretical and experimental predictions for $P_*$, defined as alignment $\alpha>0.1 $ (inset shows alignment as a function of sample size). Both theoretical and experimental results agree on $P_*\propto d$. In (c), we increase $N$ and keep $P$ and $d$ fixed, and plot the number of specialized neurons in the hidden layer. In agreement with our heuristic predictions, the number of neurons increases linearly with $\sqrt{N/d}$.  }
    \label{fig:2_layer_exp} 
\end{figure}

We now apply the heuristic principles of Sec.~\ref{sec:variational_analysis},~\ref{sec:feature_learning},~\ref{sec:kernel_prop} to derive sample complexity bounds in a few examples. In App. \ref{sec:two_layer_network}, we benchmark our method on two-layer FCNs and simple CNNs with non-overlapping patches. There we reproduce both the sample complexity exponent $P_*=d^{3/4}$ identified for CNNs in \cite{ringel_applications_2025} and further predict that $P_*=d$ for two-layer FCNs studying a Hermite-3 target as well as the scaling of the number of specializing neurons with width (see Fig. \ref{fig:2_layer_exp}). The latter is also a setting for which the prior's upper bound can be computed directly from $E(\alpha)$ using Large Deviation Theory, leading to a good match with experiment (Fig. \ref{fig:2_layer_exp} panel (a)).   

Going to what we believe is beyond the current analytical state of the art, in App. \ref{App:SoftMaxLayer}, we predict a $P_* = \sqrt{L}$, where $L$ is the context length, of a softmax attention layer learning a cubic target (see Fig. ~\ref{fig:Alignment_specialization}). Another such instance, discussed in detail below, is that of a 3-layer non-linear network learning a non-linear target. In App. ~\ref{app:classification} we show that this approach can be extended to classification tasks as well, on non-Gaussian data. In App. ~\ref{app:classification_relu_parity} we apply our heuristics to a concrete setting, of a two-layer ReLU network trained on a parity task. We find there as well are able to predict the emergent feature pattern, which qualitatively differs from erf networks. Rather than identifying the scaling number of specializing neurons, we find that there is a finite number of neurons and we are able to predict their scale.

\subsection{The Three-layer Network}
Here we consider 
three-layer FCNs given by $f(x) =  \sum_{i=1}^{N_2} a_i \sigma \big(\sum_{j=1}^{N_1} w^{(2)}_{ij} \, \sigma(w^{(1)}_j \cdot x) \big)$,
where $x \in \mathbb{R}^d$ is drawn from $\mathcal{N}(0,I_d)$. We train these networks on a polynomial target of degree $m$ given by $y(x)=He_m(w_* \cdot x)$ where $He_m$ is the $m$-th probabilist  Hermite polynomial, which is the standard polynomial choice under our choice of data measure, and $w_*\in \mathbb{R}^S$ is some normalized vector. The networks are trained via Langevin dynamics \cite{welling_bayesian_nodate}, with ridge parameter $\kappa$,  quadratic weight decay,  and standard scaling. For an extension to mean-field scaling, see App. \ref{app:mf_scaling}.  

As a starting point for our analysis, consider the simplest pattern ($q$``="{\bf GP-GP}), having two GP/lazy layers where taking an $l$'th layer to be lazy means $Q_{l,i}=K_{l-1}$ and $\mu_{l,i}=0$. Following the choice of pattern, our goal is to estimate the scaling of $\tilde{E}_q$. Examining Eq. (\ref{eq:est_E}), we find that the $\Delta_{l=1,2,i}$ contributions all cancel by our above choice of $Q$'s and $\mu$'s. The only non-zero contribution thus comes from the final layer and is given by the inverse of $\langle y, K_{2},y\rangle$. Because of lazy learning, $K_2(x,x')$ is a standard dot product FCN kernel which can be expanded as $\sum_{n=1}^{\infty} a_n (x \cdot x'/d)^n$, with $a_n=O(1)$ w.r.t. $N_{1,2},d,P$. It can then be shown that $\langle y, K_{2},y\rangle=O(d^{-m})$. Leading to $\tilde{E}_{GP-GP}=a_y \propto d^m$. 

Next, we consider a FL pattern wherein the first layer is GP distributed but the second has FL ($q$``="{\bf GP-Sp.}). Specifically, for the $l=1$ layer, we take $Q_{1,i}=K_0;\mu_{1,i}=0$ thereby nullifying again $\Delta_{1,0}$ in Eq. \ref{eq:est_E} for $\tilde{E}_q$. For the second layer, we assume $M_2$ specializing neurons (e.g., $i=1..M_2$) which fluctuate around the linear feature ($w_* \cdot x$), while others are lazy, namely $Q_{2,i>M_2}=K_{1},\mu_{2,i>M_2}=0$ and $Q_{2,i=1..M_2}=K_{1},\mu_{2,i=1..M_2}(x)=(w_* \cdot x)$. Examining $\sum_{i=1}^{N_2} \Delta_{2,i}$ in Eq. \ref{eq:est_E}, we get zero contributions from $\Delta_{2,i>M_2}=0$ and $M_2 \langle (w_* \cdot x),K_{1}^{-1},(w_* \cdot x)\rangle$ from $i=1..M_2$. As $K_1$ is again a simple FCN dot product kernel with no FL effects, normalized linear functions such as $w_* \cdot x$ have an $O(d)$ RKHS norm. We thus obtain an overall contribution to $\tilde{E}_q$ from the $l=2$ layer equal to $M_2 d$. Finally, we need to estimate $a_y=\langle y | K_2 | y \rangle^{-1}$. Note that $K_2$ is not a standard FCN kernel anymore, since $Q_2$, which contains target information, is used in its definition as the expectation of the kernel appearing in (Eq. \ref{eq:kernel_def}). According to our feature propagation rule (i), with $\Phi_*=(w_* \cdot x)$, we have $a_y = N_2/M_2$. Given $M_2$, we thus obtain a variational energy of $M_2 d+ N_2/M_2$. We next need to minimize over free parameters, namely $M_2$ leading to $M_2 = \sqrt{N_2/d}$ and finally $\tilde{E}_{{\bf GP-Sp.}}=\sqrt{N_2 d}$. Provided $N_2$ scales less than $d^{2m-1}$ ($N_2 = o(d^{2m-1})$), this pattern is favorable to ${\bf GP-GP}$. 

Finally, we consider what turns out to be the favorable pattern consisting of $M_1$ neurons specializing on $(w_* \cdot x)$ in the first layer and all neurons in the second layer specializing $He_m(w_* \cdot x)$, with a small proportionality constant $\mu_{2,i}=\pm\sqrt{\beta/N_2}$ ($q$``="{\bf Sp.-Mag.}). We refer to the second-layer pattern as magnetization. Following straightforward adaptation previous argument to this pattern, the variational energy for this pattern as well as others, for $m=3$, can be found in Table \ref{tab:fcn3_main}.
\setlength{\intextsep}{15pt}
\begin{table}[h!]
\centering
\renewcommand{\arraystretch}{1.5}
\begin{tabular}{|c|c|c|c|c|c|}
\hline
 Feature Pattern& $\Delta_1$ & $\Delta_2$ & $a_y$ & Minimizing Parameters & $\tilde{E}$ \\
\hline
\textbf{GP-GP} & $0$ & $0$ & $d^3$ & $-$ & $d^3$ \\
\hline
\textbf{GP-Sp.}& $0$ & $M_2 d$ & $\sfrac{N_2}{M_2}$ & $M_2 = \sqrt{\sfrac{N_2}{d}}$ & $\sqrt{N_2 d}$ \\
\hline
\textbf{Sp.-Mag.} & $M_1 d$ & $\sfrac{N_1 \beta}{M_1}$ & $\sfrac{N_2}{\beta}$ &
$\displaystyle \beta = \left( \sfrac{N_2^2}{N_1 d} \right)^{1/3}, \quad M_1 = \left( \sfrac{N_2 N_1}{d^2} \right)^{1/3}$ &
$\displaystyle (N_1 N_2 d)^{1/3}$ \\
\hline
\end{tabular}
\caption{Variational energy $\tilde{E}$ for different choices of feature-learning patterns in a three-layer FCN trained on $y(x)=\mathrm{He}_3(w_* \!\cdot\! x)$. The patterns shown here are (first/second layer):  
GP-GP, GP-Specialization, and Specialization-Magnetization. For each pattern, the components of the variational energy ($\Delta_1$, $\Delta_2$, $a_y$)  together with the corresponding minimizing parameters are shown.  We comment that the GP-GP pattern is favorable only for $d>N_2^5$, and otherwise FL will emerge.
 \label{tab:fcn3_main}}
\vspace{-1em}
\end{table}

In the non-GP $q$ patterns, we obtain the same scaling of $\tilde{E}_q$ in the proportional limit ($N_1 \propto N_2 \propto d $), namely, $P_*/\kappa \propto d$.  This observation is validated experimentally in Fig. \ref{fig:Alignment_specialization}, where the transition to non-zero alignment becomes sharper in the thermodynamic limit ($d\rightarrow \infty$). However, the mechanism by which this scaling is realized changes. In the specialization-magnetization pattern the sample complexity scales with $N_1^{1/3}$, therefore, it increases with $N_1$. However, under the GP-specialization pattern, sample complexity does not scale with $N_1$, making this pattern preferable. This prediction is in line with experimental results (see Figs.~\ref{fig:FCN3_phase_change} and ~\ref{fig:Alignment_specialization} panel (c)) where increasing $N_1$ causes the described change in FL patterns. Our prediction also accurately determines the scaling of the number of specializing neurons with $N_1$. 
\subsection{Softmax Attention}
\begin{figure}[h!]
    \centering
     \includegraphics[width=1.\textwidth]{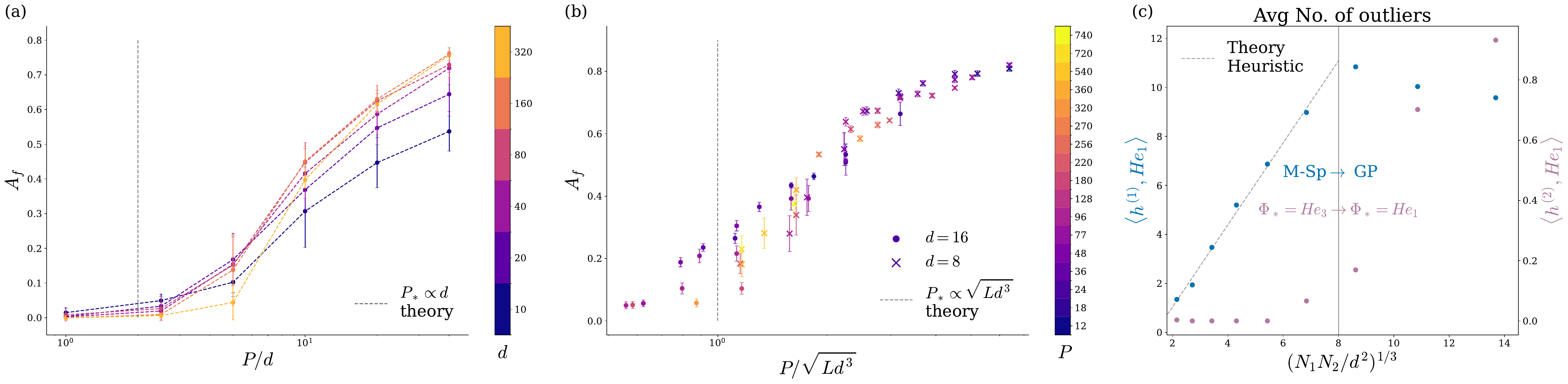}
    
    \caption{\textbf{Sample complexity:} Heuristic predictions accurately capture sample complexity in both three-layer erf FCNs and softmax attention heads, as well as feature learning scaling. Panels (a) and (b)  track network alignment against the ratio between the sample size, $P$, and the predicted sample complexity- $P/d$ for FCNs in panel (a), and $P/\sqrt{Ld^3}$ for attention heads in panel (b). In both cases, alignment collapses onto a single curve, confirming the sample complexity predictions. See Fig. ~\ref{appfig:mse} for MSE comparison and Apps. ~\ref{App:SoftMaxLayer}, ~\ref{app:3layer} for further details.    \textbf{FL patterns:} Panel (c) tracks linearly specialized neurons in the first (blue) and second (purple) layers against $N_1$ (with fixed $P, d$ and $N_2$). The number of first-layer specializing neurons initially follows the predicted $(N_1/d)^{(1/3)}$ scaling until the transition where second-layer neurons specialize on the linear feature, and the first layer approaches the GP distribution. }
    \vspace*{-0.8em}
    \label{fig:Alignment_specialization}
\end{figure}

Here, we consider an attention block of the form
\begin{align}
f(X) &= \frac{1}{\sqrt{L}} \sum_{h=1}^H\sum_{a,b=1}^L @_{ab;h}(X) (w_h \cdot x^b) , \ \ \ 
@_{ab;h}(X) = e^{[x^{a}]^{\top}A_{h}x^{b}}\big(\sum_{c=1}^{L}e^{[x^{a}]^{\top}A_{h}x^{c}}\big)^{-1},
\end{align}
where $X \in \mathbb{R}^{L \times d}$, $A \in \mathbb{R}^{d \times d}$,  and $x^a \in \mathbb{R}^d$ is the $a$-th row of $X$, and $w_h \in \mathbb{R}^d$.  
Our prior on network weights is 
$\prod_{h=1}^H \mathcal{N}(0, I_{d^2}/d^2; A_h) \, \mathcal{N}(0, I_{d}/(dH); w_h)$. The only dependence on the context length $L$ arises from the pre-factor $1/\sqrt{L}$, which ensures that for $X_i^a \sim \mathcal{N}(0,1)$, we have $f(X) = \mathcal{O}(1)$.  
The target function is given by $y(X) = \sum_{a,b} \frac{1}{\sqrt{L (L-1)}} x^a_1 x^b_2 x^b_3$, also normalized to be $\gO(1)$.  
Following our approach, we propose two patterns for this architecture: GP (or lazy learning) and specialization (where we take $A_h \sim {\cal N}(\mu \sigma_x \otimes I_{(d-2)},I_{d^2})$ for $\sigma_x=[1,0;0,1]$ and optimize over $\mu$).   
As detailed in ~\ref{App:SoftMaxLayer}, the variational energy scales as $L d^3$ for the GP pattern and as $\sqrt{H L d^3}$ for the specialization pattern, the latter thus being favorable for $H$ scaling less than $\sqrt{L d^3}$.  As shown in Fig.~\ref{fig:Alignment_specialization}, this scaling of $P$ with $L$ and $d$ indeed matches the dependence of the sample complexity on $L$.

\section{Discussion}
This paper presents a novel methodology for analyzing the scaling behavior of sample complexity, through which we can also understand how distinct learning mechanisms emerge. Its strength lies in abstracting away from fine-grained details to isolate the core principles at play. We hope such a strategy would remove barriers and expedite connections between mechanistic interpretability and first-principles scientific approaches. 

{\bf Limitations.} Notwithstanding, several avenues for improvement remain. In particular, quantifying feature propagation in more general CNNs and transformers, and addressing multi-feature interactions as in the context of superposition\cite{elhage2022superposition}. It would also be desirable to extend our heuristic to dynamics of learning, potentially drawing insights from previous work relating equilibrium and dynamical phenomena \cite{power2022grokking,rubin_grokking_2024,bahri2021explaining,nam2024exactlysolvablemodelemergence}. Since Bayesian convergence times can be slow, correctly predicting the emergence of FL in early stages of training may also be highly advantageous. Finally, in some cases, such as under mean-field scaling or vanishing ridge, overfitting effects can emerge, leaving our lower bound vacuous. 
Extending our approach to patterns that align only on the training set and incorporating effective ridge ideas \cite{Canatar2021}  is thus desirable. 

\section*{Acknowledgments}
Funded by the the Israeli Science Foundation -
374/23.


\newpage

\bibliography{iclr2026_conference}
\bibliographystyle{plainnat}

\clearpage
\appendix

\section*{Mitigating the Curse of Detail: Supplementary material}
\section{Upper Bounds on Posterior Alignment} \label{app:upper_bounds}
In this appendix, we derive bounds on the posterior probability of alignment between the network output and the target function. Our goal is to identify the scaling of the minimal sample size $P_*$  as a function of the input dimension, $d$, required for learning. The derivation proceeds by first relating posterior probabilities of trained network outputs to the prior. While the prior probability is estimated heuristically in \ref{app:E_est}, we provide here a rigorous method for computing this bound using large-deviation techniques. 

\subsection{The Posterior Bound}
Let  $f_{\Theta}$ be the output of a network with a given set of trainable weights $\Theta$.  Following standard derivations \cite{neal_bayesian_1996}, the posterior distribution of possible outputs of such a network, conditioned on training on data $X\in \mathbb{R}^{P\times d}$ with respect to target $y:\mathbb{R}^d\rightarrow\mathbb{R}$  (which we assume is normalized for simplicity), with MSE loss and ridge parameter $\kappa$, is given by
\begin{equation}
\pi\left(f\right)=\frac{1}{Z}\exp\left(-\frac{|f(X)-y(X)|^2}{2\kappa}\right)p_0\left(f\right),
\end{equation}
where $f$ and $y$ are applied row-wise to $X$, $|\cdot|$ is the $L_2$ norm on the training set, $p_0\left(f\right)$ is the prior distribution, defined by
\begin{equation}
p_0\left(f\right)=\int d\Theta \, p\left(\Theta\right)\delta\left[f-f_{\Theta}\right],
\end{equation}
$p(\Theta)$ is the prior weight distribution, and $\delta\left[f-f_{\Theta}\right]$ is the functional delta function \cite{ringel_applications_2025}.  We are interested in an upper bound on the posterior probability of achieving alignment $\geq \alpha$, where we define alignment by $A_{f}:=\langle f,y\rangle = \int d\mu_x f(x)y(x)$.  The prior  alignment density is given by
\begin{align}
p_{A_f}(\alpha) &= \int d\Theta p(\Theta) \delta(A_{f_{\Theta}}-\alpha).
\end{align}
The prior and posterior probabilities of having alignment over some threshold $\alpha$ are given by  $\Pr_{p_{0}}\left[A_{f}\geq\alpha\right]$ and $\Pr_{\pi}\left[A_{f}\geq\alpha\right]$, respectively. Since the loss is positive, we have $\Pr_{\pi}\left[A_{f}\geq\alpha\right]\leq \frac{1}{Z}\Pr_{p_{0}}\left[A_{f}\geq\alpha\right]$. We proceed by obtaining a lower bound on the partition function $Z$ --  the normalization constant of the posterior distribution -- given by
\begin{equation}
Z = \mathbb{E}_{\Theta}\left[e^{-\frac{\left|f_{\Theta}(X)-y(X)\right|^{2}}{2\kappa}}\right] := \int d\Theta \, p\left(\Theta\right)e^{-\frac{\left|f_{\Theta}(X)-y(X)\right|^{2}}{2\kappa}}.
\end{equation}
By Jensen’s inequality, we obtain
\begin{equation}
\exp\left(-\frac{\mathbb{E}_{\Theta}\left[\left|f_{\Theta}(X)-y(X)\right|^{2}\right]}{2\kappa}\right)<Z.
\end{equation}
Since $f_{\Theta}\sim\mathcal{N}\left(0,K\right)$, for the NNGP kernel $K$ on the training data \cite{Cho,neal_bayesian_1996}, it follows that $\mathbb{E}_{\Theta}\left[\left|f_{\Theta}(X)-y(X)\right|^{2}\right]=\mathrm{Tr}\left(K\right)+|y(X)|^2$. For our choice of normalization, both $\mathrm{Tr}\left(K\right)$ and $|y(X)|^2$ scale with $P$. Thus, up to an $\gO(1)$ factor $k = P^{-1} \sum_{\nu=1}^P \mathbb{E}_{p_0}[ (f(x_{\nu})-y(x_\nu))^2]$, we obtain $\mathbb{E}_{\Theta}\left[\left|f_{\Theta}(X)-y(X)\right|^{2}\right]=Pk$. Substituting this in the posterior upper bound, we obtain
\begin{align}
\Pr\nolimits_{\pi}\left[A_{f}\geq\alpha\right]<\exp\left(\frac{Pk}{2\kappa} \right)\Pr\nolimits_{p_0}\left[A_{f}\geq\alpha\right],
\end{align}
or, equivalently, 
\begin{equation}
\log\left(\Pr\nolimits_{\pi}\left[A_{f}\geq\alpha\right]\right) < Pk/\left(2\kappa\right)+\log\left(\Pr\nolimits_{p_0}\left[A_{f}\geq\alpha\right]\right).
\end{equation}
The r.h.s. is thermodynamically large in magnitude, crossing zero only briefly, since large negative values imply a vanishingly small probability of strong alignment, the zero crossing marks the threshold for learning. The minimal sample size necessary for learning is then
\begin{equation}
P_{*}=-\frac{2\kappa }{k}\log \Pr\nolimits_{p_0}\left[A_{f}\geq\alpha\right].
\label{eq:P_star_def_app}
\end{equation}
We comment that $P_*$ provides an \textit{unlearnability} bound, implying that learning cannot occur with less than $P_*$ . However, we find that $P_*$ is indeed indicative of the start of learning in the cases considered in this work. In certain settings, overfitting effects cause this bound to underestimate the true sample complexity, an effect that requires further study. In the following section, we compute the prior probability of alignment on the RHS using the Chernoff bound and large deviations theory, yielding a better estimation for $P_{*}$.

\subsection{Classification}
\label{app:classification}
In the case of a classification problem, we assume the output $f(x) \in \R^{d_o}$, where $d_o$ denotes the number of categories. The dataset is given in the form of a hot-one encoding $\gD=\left\{\left(x_\mu,y_\mu\right)\right\}_{\mu=1}^P$, where $y_\mu \in \R^{d_o}$ and $\left(y_\mu\right)_i = \delta_{c(x)i}$ for the classification target $c(x)$ taking values in $\{1,\ldots,d_o\}$. The cross-entropy loss for this setting is
\begin{align*}
    \gL(f;\gD) & = - \sum_{\mu=1}^P \left[\log\left(\frac{e^{-f_{c(x_\mu)}(x_\mu)}}{\sum_{j=1}^{d_o} e^{-f_j(x_\mu)}}\right)\right] \\
    & = \sum_{\mu=1}^P \left[ f_{c(x_\mu)}(x_\mu) + \log\left(\sum_{j=1}^{d_o} e^{-f_j(x_\mu)}\right) \right] \geq 0
\end{align*}
The derivation 
\begin{equation}
\pi\left(f\right)=\frac{1}{Z}\exp\left(-\frac{\gL(f;\gD)}{2\kappa}\right)p_0\left(f\right),
\end{equation}
follows that of the MSE loss. We similarly have $\Pr_{\pi}\left[A_{f}\geq\alpha\right]\leq \frac{1}{Z}\Pr_{p_{0}}\left[A_{f}\geq\alpha\right]$, since the cross-entropy loss is also positive. The use of Jensen’s inequality does not depend on $\gL(f;\gD)$, so we only need to compute $\E_{\Theta}\left[\gL(f;\gD)\right]$. 

By \cite{Cho,neal_bayesian_1996}, we have $f_j \sim \mathcal{GP}\left(0,K_j\right)$ for the NNGP kernel $K_j$. Since we only care about the values of $f$ at $\gD$, we consider $f_j \sim \gN(0,K_j)$, $K_j \in \R^{P \times P}$, $(K_j)_{\mu\nu} = K_j(x_\mu,x_\nu)$, where we overloaded the notation $f_j$ and $K_j$ for simplicity. 
\begin{align*}
    \E_{\Theta}\left[\gL(f;\gD)\right] &= 
    \E_{f \sim \prod_{j=1}^{d_o} \gN(0,K_j)}\left[\gL(f;\gD)\right] \\
    & = \E_{f \sim \prod_{j=1}^{d_o} \gN(0,K_j)} \left[ \sum_{\mu=1}^P \left[ f_{c(x_\mu)}(x_\mu) + \log\left(\sum_{j=1}^{d_o} e^{-f_j(x_\mu)}\right) \right] \right] \\
    & = \sum_{\mu=1}^P \E_{f \sim \prod_{j=1}^{d_o} \gN(0,K_j)} \left[ f_{c(x_\mu)}(x_\mu) + \log\left(\sum_{j=1}^{d_o} e^{-f_j(x_\mu)}\right) \right]
\end{align*}
Since each summand depends on $x_\mu$ alone, we can replace $f \sim \prod_{j=1}^{d_o} \gN(0,K_j)$ by $f_\mu \sim \gN(0,\Sigma_\mu)$ where $\left(\Sigma_\mu\right)_{ij} = \delta_{ij} \sigma_{\mu,j}^{2}$, $i,j \in \{1,\ldots,d_o\}$, $\sigma_{\mu,j}^{2} = K_j(x_\mu,x_\mu)$. We will also denote $c_\mu:=c(x_\mu)$.
\begin{align*}
    \E_{\Theta}\left[\gL(f;\gD)\right] & = \sum_{\mu=1}^P \E_{f_\mu \sim \gN(0,\Sigma_\mu)} \left[ \left(f_\mu\right)_{c_\mu} + \log\left(\sum_{j=1}^{d_o} e^{-\left(f_\mu\right)_j}\right) \right] \\
    & = \sum_{\mu=1}^P \E_{f_\mu \sim \gN(0,\Sigma_\mu)} \left[ \log\left(\sum_{j=1}^{d_o} e^{-\left(f_\mu\right)_j}\right) \right]
\end{align*}
Let $\psi(f_\mu) := \log \left(\sum_{j=1}^{d_o} e^{-\left(f_\mu\right)_j}\right)$, so we end up with
\begin{align*}
    \E_{\Theta}\left[\gL(f;\gD)\right] & = \sum_{\mu=1}^P \E_{f_\mu \sim \gN(0,\Sigma_\mu)} \left[ \psi(f_\mu) \right]
\end{align*}
Let us concentrate on a single summand and drop $\mu$ from our notation for the moment. By Wick's Theorem, we have 
\begin{align*}
    \E_{f \sim \gN(0,\Sigma)} \left[ \psi(f) \right] &= e^{\frac{1}{2} \sum_{j=1}^{d_o} \sigma_j^{2} \frac{\partial_j^2}{\partial f_j^2}} \left.\psi(f)\right|_{f=0} \\
    &= \psi(0) + \frac{1}{2} \sum_{j=1}^{d_o} \sigma_j^{2} \frac{\partial_j^2}{\partial f_j^2} \left.\psi(f)\right|_{f=0} + \mathrm{H.O.T} \\
    &= \log(d_o) + \frac{d_o-1}{d_o^2} \sum_{j=1}^{d_o} \sigma_j^{2} + \mathrm{H.O.T}
\end{align*}
Bringing the sum over datapoints back and assuming $\mathrm{Tr}\left(K_j\right)$ scales as $P$, we have
\begin{align*}
    \E_{f \sim \prod_{j=1}^{d_o} \gN(0,K_j)}\left[\gL(f;\gD)\right] & = P\log(d_o) + \underbrace{\frac{1}{d_o} \sum_{j=1}^{d_o} \mathrm{Tr}(K_j)}_{\gO(P)} + \gO\left(\frac{P}{d_o}\right)
\end{align*}
where the subleading correction $\gO\left({P}/{d_o}\right)$ comes from an $\gO(1/d_o)$ contribution of the $2n$-th derivative for $n\geq 2$ combined with summation over $d_o$ classes and $P$ stems from the summation over datapoints. Thus, for a classification task, we obtain
\begin{align}
\Pr\nolimits_{\pi}\left[A_{f}\geq\alpha\right]<\exp\left(\frac{P\log(d_o)}{2\kappa} \right)\Pr\nolimits_{p_0}\left[A_{f}\geq\alpha\right],
\end{align}
Consequently, Eq.~\ref{eq:P_star_def_app} applies with the necessary adjustment, replacing $k$ with $\log(d_o)$. 

Our use of Wick’s Theorem to compute the disorder–averaged log-partition function over classes is closely related to the annealed and quenched free-energy calculations that appear in the statistical-physics literature on disordered systems. In that language, for fixed data $\gD$ and random parameters 
$\Theta$, the quantity $-\gL(f_\Theta;\gD)$ plays the role of a random energy, and $-\E_\Theta[\gL(f_\Theta;\gD)]$ is the quenched free energy of a disordered system with $d_o$ states. Our leading term $P\log(d_o)$ is then the purely 
entropic contribution coming from $d_o$ equiprobable states 
per datapoint, in direct analogy with the $N \log(2)$ term in Derrida's random–energy model \cite{Derrida1980REM} for a system with $2^N$ configurations. In the spin-glass literature, one often replaces the quenched average $\E[\log(Z)]$ by the annealed approximation $\log \E[Z]$, which follows from Jensen's 
inequality for $\log$ and is accurate in the replica-symmetric/high-temperature phase. Our calculation stays on the quenched side and is elementary (Gaussian integrals plus Wick's theorem),  but the resulting scaling structure -- an entropic $\log(d_o)$ term with subleading corrections controlled by the covariance of the random energies -- is entirely consistent with this broader statistical-mechanics viewpoint.

\subsection{The Chernoff Upper Bound}
\label{app:upper_bound_chern}
We refine the posterior bound using the Chernoff inequality. This allows us to estimate the probability of achieving alignment $\geq\alpha$ under the prior distribution, leading to an explicit expression for the corresponding sample complexity (\ref{eq:P_star_def_app}). From the Chernhoff bound, we get
\begin{equation}
\label{eq:chern_bound_def}
\Pr\nolimits_{p_0}\left[A_{f}\geq\alpha\right] \leq\inf_{t>0}e^{-t\alpha}\mathbb{E}_{\Theta}\left[e^{tA_{f}}\right] =: e^{ -E\left(\alpha\right)}.
\end{equation}
Substituting in (\ref{eq:P_star_def_app}), we obtain a new estimate 
\begin{equation}
\label{eq:chern_P_star}
P_{*}=\frac{2\kappa }{k}E\left(\alpha\right).
\end{equation}
Our aim is to better estimate $E(\alpha)$. Following standard complex analysis derivation, we have
\begin{align}
-\log(\Pr\nolimits_{p_0}\left[A_{f}\geq\alpha\right]) \geq E(\alpha) &:=-\log(\inf_{t>0}\mathbb{E}_{\Theta}\left[\exp\left(t\left(A_{f_{\Theta}}-\alpha\right)\right)\right] ) \nonumber \\
& = -\log (\ext_{t\in\mathbb{C}}\mathbb{E}_{\Theta}\left[\exp\left(it\left(A_{f_{\Theta}}-\alpha\right)\right)\right]),
\end{align}
where $\ext$ denotes the extremum with respect to a complex $t$. Since we are in the rare event regime, we can apply a saddle point approximation to the Fourier transform of $p_{A_f}(\alpha)$ 
\begin{equation}
p_{A_f}\!\left(\alpha\right)=\frac{1}{2\pi}\int dt\mathbb{E}_{\Theta}\left[\exp\left(it\left(A_{f_{\Theta}}-\alpha\right)\right)\right]\approx\ext_{t\in\mathbb{C}}\mathbb{E}_{\Theta}\left[\exp\left(it\left(A_{f_{\Theta}}-\alpha\right)\right)\right].
\end{equation}
Thus, (\ref{eq:chern_bound_def}) can be reduced to 
\begin{equation}
-\log\Pr\nolimits_{p_0}\left[A_{f}\geq\alpha\right] \geq -\log  p_{A_f}(\alpha)\approx E(\alpha).
\end{equation}
This definition of the upper bound allows for furhter simplifications.

\section{An Explicit Equation for the LDT Bound in Fully Connected Networks}\label{app:upper_bound_chern_FCN}

To make our LDT analysis concrete, we now focus on FCNs. This can be trivially adapted to other architectures, however, such as CNNs and transformers. Here, we provide an explicit expression for the minimal $t_*$ in Eq. ~\ref{eq:chern_bound_def}. The resulting formula involves high-dimensional, non-Gaussian integrals and is therefore highly computationally intensive. We also work out the two-layer case explicitly, where we show that the solution can be reduced to one-dimensional integrals. 

\subsection{Setup}

We begin by specifying the architecture and notation used throughout this section. Focusing on FCNs of depth $L>1$, we define the forward pass and parameterization explicitly:
\begin{align*}
f\left(x\right) & =\sum_{i=1}^{N_{L-1}}w_{i}^{L}\sigma\left(h_{i}^{L-1}\left(x\right)\right)\\
h_{i}^{l}\left(x\right) & = \sum_{j=1}^{N_{l-1}} W_{ij}^{l}\sigma\left(h_{j}^{l-1}\left(x\right)\right)\ \ l=2,...,L-1,\ \ i=1,...,N_l\\
h^{1}_i\left(x\right) & =[W^{1}x]_i,\ \ i=1,...,N_1
\end{align*}
We train the network using Langevin dynamics~\cite{Welling2011} with MSE loss on a target function $y$. The quadratic weight decay for each layer $l$ is set to $\kappa N_{l-1}/\sigma_l^2$ where $\kappa$ is the ridge parameter (corresponding to the Langevin dynamics temperature, $T$ \cite{ringel_applications_2025}) and $N_0=d$ is the input dimension. We define $\sigma_L^2,\kappa\sim\mathcal{O}(1/\chi)$, where we refer to  $\chi$  as the mean-field (MF) scale. This choice of weight decay results in a Gaussian prior distribution for the weights given by $W_{i,j}^{l}\sim\mathcal{N}(0,\sigma_l^2/N_{l-1})$ for $i=1,..,N_{l}$ , $j=1,...,N_{l-1}$. 

\subsection{Derivation}

In the following, we will derive an explicit equation for $t_*$ that minimizes the upper bound for $\Pr_{p_0}[A_f\geq\alpha]$ in (\ref{eq:chern_bound_def}). We will use $\Theta^{l}$ to denote the weights up to (and including) layer $l$, in particular, $\Theta^L\equiv\Theta$. Utilizing this notation, we can separate the readout weights,
\begin{multline}
\label{eq:mf_seperated_dist}
\mathbb{E}_\Theta\left[e^{t\left(A_f -\alpha\right)}\right] \propto \\ e^{-t\alpha}\int\prod_{i=1}^{N_{L-1}}dw_{i}^{L}d\Theta^{L-1}p\left(\Theta^{L-1}\right)\exp\left(-\frac{N_{L-1}}{2\sigma_{L}^{2}}\left(w_{i}^{L}\right)^{2}+t\sum_{i=1}^{N_{L}}w_{i}^{L}\left\langle \sigma\left(h_{i}^{L-1}\left(x\right)\right), y\left(x\right)\right\rangle \right).    \end{multline}
We integrate out the weights of the final layer to get
\begin{align}
\mathbb{E}_\Theta\left[e^{t\left(A_{f} -\alpha\right)}\right]   & \propto e^{-t\alpha}\int d\Theta^{L-1}p\left(\Theta^{L-1}\right)\exp\left(\frac{\sigma_{L}^{2}}{2N_{L-1}}t^{2}\underbrace{\sum_{i=1}^{N_{L-1}}\left\langle \sigma\left(h_{i}^{L-1}\left(x\right)\right), y\left(x\right)\right\rangle ^{2}}_{A_\sigma^2\left(\Theta^{L-1}\right)}\right).
\label{eq:readout_integration_LDT}
\end{align}
If $\alpha$ is such that it is smaller than the upper bound of $A_f$, which occurs naturally for unbounded $f$  as in our case where the readout weights are unbounded, then $t_*\in (0,\infty)$. Thus, $t_*$  is determined by minimizing with respect to $t$, and we obtain
\begin{align*}
0 
& = \left.\frac{\partial\mathbb{E}_\Theta\left[e^{t\left(A_{f}-\alpha\right)}\right]}{\partial t}\right|_{t=t_*} \\
& = -\alpha e^{-t\alpha}\int d\Theta^{L-1}p\left(\Theta^{L-1}\right)\exp\left(\frac{\sigma_{L}^{2}}{2N_{L-1}}\sum_{i=1}^{N_{L-1}}t^{2}A_{\sigma,i}^{2}\left(\Theta^{L-1}\right)\right) \\
& \ \ \ \ +\left.\sum_{i=1}^{N_{L-1}}\frac{\sigma_{L}^{2}}{N_{L-1}}te^{-t\alpha}\int d\Theta^{L-1}p\left(\Theta^{L-1}\right)A_{\sigma,i}^{2}\left(\Theta^{L-1}\right)\exp\left(\frac{\sigma_{L}^{2}}{2N_{L-1}}\sum_{i=1}^{N_{L-1}}t^{2}A_{\sigma,i}^{2}\left(\Theta^{L-1}\right)\right)\right|_{t=t_*},
\end{align*}
where we have $A_{\sigma,i}^{2}\left(\Theta^{L-1}\right):=\left\langle \sigma\left(h_{i}^{L-1}\right), y\right\rangle^2$. Rearranging, we obtain an implicit equation for $t_*$
\begin{equation}
t_*=\frac{\alpha N_{L}\int d\Theta^{L-1}p\left(\Theta^{L-1}\right) \exp\left(\frac{\sigma_{L}^{2}}{2N_{L-1}}t_*^{2} A_\sigma^2\left(\Theta^{L-1}\right)\right)}{\sigma_{L}^{2}\int d\Theta^{L-1} p\left(\Theta^{L-1}\right) A_\sigma^2\left(\Theta^{L-1}\right) \exp\left(\frac{\sigma_{L}^{2}}{2N_{L-1}}t_*^{2}A_\sigma^2\left(\Theta^{L-1}\right)\right)},
\end{equation}
which takes a simpler form, using the notation $\mathbb{E}_{\Theta^{L-1}}\left[\left(\cdots\right)\right]=\int d\Theta^{L-1}p\left(\Theta^{L-1}\right)\left(\cdots\right)$,
\begin{equation}
t_{*}=\frac{\alpha N_{L-1} \mathbb{E}_{\Theta^{L-1}}\left[\exp\left(\frac{\sigma_{L}^{2}
}{2N_{L-1}}t_*^{2} A_\sigma^2\left(\Theta^{L-1}\right)\right)\right]}{\sigma_{L}^{2}  \mathbb{E}_{\Theta^{L-1}} \left[A_\sigma^2\left(\Theta^{L-1}\right) \exp\left(\frac{\sigma_{L}^{2}}{2N_{L-1}}t_{*}^{2}A_\sigma^2\left(\Theta^{L-1}\right)\right)\right]}.
\label{eq:min_t}
\end{equation}

\subsection{Exact LDT Saddle-point Computation for a Two-layer Network}

To illustrate the general framework in a more tractable setting, we analyze a simple two-layer FCN. In this case, the derivation simplifies substantially, allowing us to obtain explicit analytical expressions assuming $x \sim \gN(0,I_d)$. Concretely, consider the setting
\begin{equation*}
    y\left(x\right)=\hat{H}e_{3}\left(w_{*}\cdot x\right),\quad\sigma=\text{erf},\quad L=2
\end{equation*}
where $w_*\in\mathbb{R}^d$ is some normalized vector, and  $\hat{H}e_3$ is the normalized third probabilist  Hermite polynomial. Since this is a simple two-layer network, we need only consider the weights of the hidden layer themselves, and not the pre-activations.
For brevity, we also drop layer indexing ($W \equiv W^1 = \Theta^1$, $N \equiv N_1$, $d = N_0$), since there is only one relevant layer, and take $\sigma_{1}^{2},\sigma_{2}^{2}=1$, in particular, we assume standard scaling. Following (\ref{eq:min_t}), we obtain 
\begin{align*}
    t_{*}=\frac{\alpha N \mathbb{E}_{W}\left[\exp\left(\frac{1
    }{2N}t_*^{2} A_\sigma^2\left(W\right)\right)\right]}{\mathbb{E}_{W} \left[A_\sigma^2\left(W\right) \exp\left(\frac{1}{2N}t_{*}^{2}A_\sigma^2\left(W\right)\right)\right]}
\end{align*}
where $A_{\sigma}^{2}(W) = \sum_{i=1}^N \left\langle \sigma\left(w_{i}\cdot x\right), y\left(x\right)\right\rangle^2$, using $w_i \in \sR^d$ for the $i$-th row of $W$.
Since $W$ is i.i.d. in the different neuron indices,
the above equation becomes
\begin{align}
t_* & =\frac{\alpha\int d^{d}w\exp\left(-\frac{d}{2}\left|w\right|^{2}+\frac{1}{2N}t_*^{2}A_{\sigma}^{2}\left(w\right)\right)}{\int d^{d}wA_{\sigma}^{2}\left(w\right)\exp\left(-\frac{d}{2}\left|w\right|^{2}+\frac{1}{2N}t_*^{2}A_{\sigma}^{2}\left(w\right)\right)}
\label{eq:t_min_2_layer}
\end{align}
where $A_{\sigma}^{2}\left(w\right)=\left\langle \sigma\left(w\cdot x\right), y\left(x\right)\right\rangle^2$, $w \in \sR^d$, $w \in \gN\left(0,\frac{1}{d}I_d\right)$. 
This inner product can be calculated explicitly. For a general (normalized) probabilist Hermite polynomial, using the Hermite Rodrigues formula together with high-order integration by parts (as a generalization of Stein's identity) produces 
\begin{align*}
\mathbb{E}_{x \sim \mathcal{N}(0, I_d)} 
\left[
\hat{H}e_{n}\left(w_*^{\top} x\right)\,
\sigma\left(w^{\top} x\right)
\right]
& = \frac{\left(w_*^{\top} w\right)^{n}}{n!}
\,\mathbb{E}_{x \sim \mathcal{N}(0, I_d)}\left[
\sigma^{(n)}\left(w^{\top} x\right)
\right] \\
& = \frac{\left(w_*^{\top} w\right)^{n}}{n!}
\,\mathbb{E}_{t \sim \mathcal{N}(0, |w|^2)}\left[
\sigma^{(n)}(t)
\right]
\end{align*}
assuming $|w_*|=1$, where $\sigma^{(n)}$ stands for the the $n$-th derivative of $\sigma$. 
From here on, we simply use erf's third derivative, giving us straightforward Gaussian integrals to calculate that produce
\begin{equation}
\left\langle \text{erf}\left(w\cdot x\right), y\left(x\right)\right\rangle ^{2}=\frac{4}{9\pi}\left(\frac{w\cdot w_{*}}{\sqrt{1+2\left|w\right|^{2}}}\right)^{6}\label{eq:A_sigma}
\end{equation}
Next, assuming that all components of $w$ perpendicular to $w_{*}$ are weakly fluctuating, we can approximate 

\begin{equation*}
\frac{w\cdot w_{*}}{\sqrt{1+2\left|w\right|^{2}}} \approx \frac{w\cdot w_{*}}{\sqrt{1+2\left(w_{*}\cdot w\right)^{2} + 2\sum_{\perp}\mathbb{E}_{w\cdot w_{\perp}}\left[\left(w\cdot w_{\perp}\right)^{2}\right]}} \approx \frac{w\cdot w_{*}}{\sqrt{3+2\left(w_{*}\cdot w\right)^{2}}},
\end{equation*}
where $\sum_{\perp}$ denotes a sum over an orthonormal basis in weight space, $w_{\perp}$, which are orthogonal to $w_*$. 
Using $\E_{w \sim \gN(0,\frac{1}{d}I_d)} \left[g(w \cdot w_*)\right] = \E_{\beta \sim \gN(0,\frac{1}{d})} \left[g(\beta)\right]$ for $\beta=w \cdot w_*$ with a unit vector $w_*$, and for any function $g$, we can apply the above approximation to (\ref{eq:t_min_2_layer}) to get
\begin{align*}
t_* & \approx \frac{9\alpha\pi \E_{\beta\sim\gN(0,\frac{1}{d})} \left[\exp\left(\frac{2}{9\pi N}t_*^{2} \left(\frac{\beta}{\sqrt{3+2\beta^{2}}}\right)^6 \right)\right]}{4\E_{\beta\sim\gN(0,\frac{1}{d})} \left[\left(\frac{\beta}{\sqrt{3+2\beta^{2}}}\right)^6 \exp\left(\frac{2}{9\pi N}t_*^{2} \left(\frac{\beta}{\sqrt{3+2\beta^{2}}}\right)^6 \right)\right]}
\end{align*}
If we then denote 
\begin{align}
\label{eq:S_beta}
    S\left(\beta;t_*\right)=\beta^{2}-\frac{4t_*^{2}}{9\pi N d}\left(\frac{\beta}{\sqrt{3+2\beta^{2}}}\right)^{6},
\end{align}
we end up with
\begin{equation}
t_* \approx \frac{9\alpha\pi}{4}\cdot\frac{\int d\beta\exp\left(-\frac{d}{2}S\left(\beta;t_*\right)\right)}{\int d\beta\left(\frac{\beta}{\sqrt{3+2\beta^{2}}}\right)^{6}\exp\left(-\frac{d}{2}S\left(\beta;t_*\right)\right)},
\label{eq:2_layer_explicit}
\end{equation}
The $t_*$, which (approximately) solves (\ref{eq:2_layer_explicit}), can then be used to compute the scaling of $P_*$. 

Following the derivation for the upper bound (\ref{eq:chern_P_star}), recall that we have
\begin{equation}
P_{*}\propto-2\kappa\log\left(e^{-t_*\alpha}\mathbb{E}_{\Theta}\left[e^{t_*A_{f}}\right]\right)\label{eq:sample_comp_scale_FCN2}
\end{equation}
for $k\approx 1$.
Following (\ref{eq:readout_integration_LDT}), we integrate out the readout weights and utilize our calculation, culminating in (\ref{eq:2_layer_explicit}), to get
\begin{align}
\mathbb{E}_{\Theta}\left[e^{t_*A_{f}}\right] &= \mathbb{E}_{W}\left[e^{\frac{1}{2N}\sum_{i=1}^{N}t_*^{2}A_{\sigma,i}^{2}\left(W\right)}\right] \nonumber \\
&= \prod_{i=1}^N \mathbb{E}_{w_i \sim \gN(0,\frac{1}{d}I_d}) \left[e^{\frac{1}{2N} t_*^{2} A_{\sigma}^{2}(w_i)}\right] \nonumber \\
&\approx \left[ \int d\beta\exp\left(-\frac{d}{2}S\left(\beta;t_*\right)\right) \right]^N.\label{eq:2_layer_exact_partition}
\end{align}

Finally, we arrive at the following expression for sample complexity
\begin{equation}
\label{eq:P_*_2_layer}
P_{*} \propto 2\kappa \left[ t_* \alpha - N \log \left(\int d\beta\exp\left(-\frac{d}{2}S\left(\beta;t_*\right)\right)\right)\right]
\end{equation}
To evaluate it, we numerically solve for $t_*$ in (\ref{eq:2_layer_explicit}). The resulting $t_*$ is then used to numerically evaluate the one-dimensional integral in (\ref{eq:2_layer_exact_partition}) over $\beta=w\cdot w_*$. This produces the sample complexity scale in (\ref{eq:P_*_2_layer}). These computational results are shown in Fig.~\ref{fig:2_layer_exp}, including predictions both for network output and weight distribution.

We can gain further insight by interpreting (\ref{eq:2_layer_exact_partition}) as the partition function (i.e., the normalization factor) of a $t_*$-dependent, non-Gaussian distribution. This distribution can be understood as representing the required weight configuration for achieving good alignment. While good alignment is improbable under the prior, this new distribution makes it attainable. If this distribution represents the optimal path to alignment within the prior, it is reasonable to infer that it will correspond to the FL patterns in the posterior. Indeed, as illustrated in Fig.~\ref{fig:2_layer_exp}, this is precisely the case, as the posterior weight distribution is closely approximated by the prior alignment distribution we predict. It is this observation that motivates the reasoning in the following sections, by guessing various FL patterns that result in good alignment in the prior, we can deduce the actual distribution in the posterior. We can compare different guesses and determine which of the patterns is most likely to emerge. 

\section{Variational Estimate of \texorpdfstring{$E(\alpha)$}{E(alpha)}}
\label{app:E_est}

As the explicit expression for the energy, $E(\alpha)=-\log p_{A_f}(\alpha)$, is intractable, we turn to computing a variational estimate for this quantity.  

\subsection{Distribution Analysis in Pre-activation space}
\label{app:layerwise_seperation}
We first rewrite $p_{A_f}$ as a distribution on the pre-activations, rather than the network weights. Following \ref{app:upper_bound_chern}, the upper bound for the prior probability is given by 
\begin{equation}    \Pr\nolimits_{p_{0}}\left[A_{f}\geq\alpha\right]\leq p_{A_f}(\alpha):= \int d\Theta p\left(\Theta\right)\delta\left(A_{f_{\Theta}}-\alpha\right).
\end{equation} 
Replacing the delta function with its Fourier representation, $p_{A_f}(\alpha) = \frac{1}{2\pi} \int dt e^{-it\alpha} \hat{p}_{A_f}(t)$, $\hat{p}_{A_f}(t) = \E_\Theta \left[\exp\left(itA_f\right)\right]$, we obtain
\begin{equation}
    p_{A_f}(\alpha) = \frac{1}{2\pi}\int dt\int d\Theta p\left(\Theta\right)\exp\left(it\left(\sum_{i=1}^{N_{L-1}}w_{i}^{L}\left\langle y,\sigma\left(h_{i}^{L-1}\right) \right\rangle -\alpha\right)\right).
\end{equation}
We can simplify this integral by integrating out the readout weights in the readout layer, as done in  (\ref{eq:readout_integration_LDT}), which results in
\begin{equation}
p_{A_f}(\alpha) = \frac{1}{2\pi}\int dt \int d\Theta^{L-1}p\left(\Theta^{L-1}\right)\exp\left(-\frac{\sigma_L^2}{2N_{L-1}}\sum _{i=1}^{N_{L-1}}t^{2}\left\langle y,\sigma (h_{i}^{L-1})\right\rangle ^{2} -it\alpha\right)
\end{equation}
By comparison, in (\ref{eq:mf_seperated_dist})  $t$ was real, accounting for the difference in sign. Integrating out $t$ we obtain
\begin{align}
& p_{A_f}(\alpha) = \\
& = \frac{1}{\sqrt{2\pi}} \int d\Theta^{L-1}p\left(\Theta^{L-1}\right)\exp\left(-\frac{\alpha^{2}}{\frac{2\sigma_{L}^{2}}{N_{L-1}}\sum_{i=1}^{N_{L-1}}\left\langle y,\sigma\left(h_{i}^{L-1}\right)\right\rangle ^{2}}+\frac{1}{2}\log\frac{1}{\frac{\sigma_{L}^{2}}{N_{L-1}}\sum_{i=1}^{N_{L-1}}\left\langle y,\sigma\left(h_{i}^{L-1}\right)\right\rangle ^{2}}\right)\nonumber\\
& =\int d\Theta^{L-1}p\left(\Theta^{L-1}\right)\mathcal{N}(\alpha|0,\kappa_{A}(h^{L-1}(\Theta^{L-1}))),\nonumber 
\end{align}
where the variance $\kappa_{A}$ is  given by 
\begin{equation}\kappa_{A}\left(h^{L-1}\left(\Theta^{L-1}\right)\right)=\frac{\sigma_{L}^{2}}{N_{L-1}}\sum_{i=1}^{N_{L-1}}\left\langle y,\sigma\left(h_{i}^{L-1}\left(\Theta^{L-1}\right)\right)\right\rangle^{2}
\end{equation}
and $h_i^{L-1}$ are the final layer's pre-activations, which are themselves determined by the weights of the first $L-1$ layers of the network \(\Theta^{L-1}\).  We now perform a change of variables by enforcing pre-activations through a delta function and then rewriting it in terms of its Fourier representation. We then obtain
\begin{equation}
\begin{aligned}
p_{A_f}(\alpha)
&= \int \mathcal{D}h \int \mathrm{d}W \int \mathcal{D}m\,
   \exp\!\left(-\sum_{l=1}^{L-1}\sum_{i=1}^{N_l}\sum_{j=1}^{N_{l-1}}
      \frac{N_{l-1}}{2\sigma_l^{2}}\left( W_{ij}^{l}\right)^{2}\right) \\
&\quad\times
   \exp\!\left(i \sum_{l=1}^{L-1}\sum_{i=1}^{N_l} \left[\Big\langle m_{i}^l, h_i^{l} \Big\rangle
      - \Big\langle m_{i}^l, \sum_{j=1}^{N_{l-1}} W_{ij}^{l}\,
        \sigma(h_{j}^{l-1}) \Big\rangle \right]\right) \mathcal{N}(\alpha|0,\kappa_{A}\left(h^{L-1}\right)) \\
&=: \int \mathcal{D}h\, \tilde{p}(h)\,\mathcal{N}\!\big(\alpha \mid 0,\kappa_A(h^{L-1})\big).
\end{aligned}
\end{equation}
Here $\mathcal{D}h$ denotes the path integral over the pre-activation functions $h$ for all layers and neurons, and 
$\mathcal{D}m$ is the path integral over its Fourier conjugate variables (see \cite[Sec.~3.1]{ringel_applications_2025} for further details). We further specify that the integration measures are normalized to include the appropriate powers of $\pi$. When $l=1$ we replace $\sigma(h_{j}^{l-1})$ with $x$. We next integrate out the weights $W$ to obtain 
\begin{align}
\tilde{p}\left(h\right)=\,\int \mathcal{D}m\exp\!\left(-\sum_{l=1}^{L-1}\sum_{i=1}^{N_{l}}\sum_{j=1}^{N_{l-1}}\frac{\sigma_{l}^{2}}{2N_{l-1}}\left\langle m_{i}^l,\sigma(h_{j}^{l-1})\right\rangle ^{2}+i\sum_{l=1}^{L-1}\sum_{i=1}^{N_{l}}\left\langle m_{i}^l,h_{i}^{l}\right\rangle \right).
\label{eq:pre_mf}
\end{align}
Finally, we integrate out the auxiliary fields $m_{i}^l$ to obtain
\begin{align}
\label{Eq:InterLayerActions}
\tilde{p}\left(h\right)=\exp\!\left(-\frac{1}{2}\sum_{l=1}^{L-1}\sum_{i=1}^{N_{l}}\langle h_{i}^{l},[\tilde{K}_{l-1}]^{-1},h_{i}^{l}\rangle-\frac{1}{2}\Tr\log(\tilde{K}_{l-1})\right).
\end{align}
where the integral operator 
\begin{align*}
    \tilde{K}_{l-1}(x,x') = \frac{\sigma_l^2}{N_{l-1}} \sum_{j=1}^{N_{l-1}} \sigma(h_j^{l-1}(x))\sigma(h_j^{l-1}(x'))
\end{align*}
can be understood as the fluctuating $h^{l-1}$-dependent kernel and $[\tilde{K}_{l-1}]^{-1}$ is its inverse w.r.t. the measure $d\mu_x$. The above is an exact rewriting of the probability distribution showing that, conditioned on previous layers, each hidden layer has Gaussian pre-activations. Inspired by standard statistical mechanics notation, we define the for $\tilde{p}$ the Hamiltonian $H_{\tilde{p}}(h)=\frac{1}{2}\sum_{l=1}^{L-1}\sum_{i=1}^{N_{l}}\langle h_{i}^{l},[\tilde{K}_{l-1}]^{-1},h_{i}^{l}\rangle$ and the fluctuating "partition function" $Z_{\tilde{p}}$ such that $\log Z_{\tilde{p}}(h)=\frac{1}{2}\sum_{l=1}^{L-1}\Tr\log(\tilde{K}_{l-1})$. Thus, we can write
$\tilde{p}(h) = Z_{\tilde{p}}^{-1}(h) \exp[-H_{\tilde{p}}(h)]$.

\subsection{Variational approximation}
\label{app:variation_est_intro}
Turning back to the alignment density $p_{A_f}$, we can substitute the result from the previous section so that
\begin{equation}
    p_{A_{f}}(\alpha)=\int\mathcal{D}h\exp \left(-\underbrace{\left(H_{\tilde{p}}\left(h\right)+\frac{\alpha^{2}}{2\kappa_{A}\left(h^{L-1}\right)}\right)}_{=:-H_{p,\alpha}\left(h\right)}- \underbrace{\left(\frac{1}{2}\log\kappa_{A}(h^{L-1})+\log Z_{\tilde{p}}(h)\right)}_{=:\log Z_{A_f}(h)}\right) 
\end{equation}
We are interested in computing $E(\alpha)$, which is given by 
\begin{equation}
    E(\alpha)=-\log p_{A_{f}}(\alpha)=-\log\int\mathcal{D}h\exp\left(-H_{p,\alpha}\left(h\right)\right)/Z_{A_f}(h).
\end{equation}
Since the integral over \(h\) is generally intractable, we turn to a variational approximation. The goal of this estimate is to find a simpler, tractable distribution $q_{\alpha}(h)$ that is in some sense "close" to the true pre-activation density $\exp(-H_{p,\alpha}(h))$ per alpha. We thus define our variational distribution in a similar Hamiltonian
\begin{equation}
    q_{\alpha}(h) = \frac{1}{Z_{q,\alpha}} \exp(-H_{q,\alpha}(h)),
\end{equation}
where $H_{q,\alpha}$  satisfies $\min_hH_{q,\alpha}(h)=0$, and  $Z_{q,\alpha}$ is the partition function given by $Z_{q,\alpha}=\int\mathcal{D}h\exp\left(-H_{q,\alpha}\left(h\right)\right)$. This choice results in a unique definition of $H_{q,\alpha},\ Z_{q,\alpha}$ for each distribution $q_{\alpha}$. The Feynman-Bogoliubov inequality provides a rigorous definition of closeness between $\exp(-H_{p,\alpha}(h))$ and $q_\alpha(h)$. This inequality establishes an upper bound on $E(\alpha)$ which depends on the choice of $q$. The Feynman-Bogoliubov inequality \cite{bogolubov_introduction_2009} states that  for any $q$
\begin{equation}
    -\log\left(\int\mathcal{D}he^{-H_{p,\alpha}\left(h\right)}/Z_{A_f}(h)\right)\leq-\log Z_{q,\alpha}+\mathbb{E}_{h\sim q_{\alpha}}\left[H_{p,\alpha}(h)-H_{q,\alpha}(h)+\log Z_{A_f}(h)\right]
\end{equation}
where  \(\mathbb{E}_{h\sim q_\alpha}[\cdot]\) is the expectation over \(h\) with respect to the distribution \(q_\alpha\). Recalling that  $E(\alpha) = -\log p_{A_f}(\alpha)$, this inequality can be used to simplify the bound
\begin{equation}
    E(\alpha) \leq \mathbb{E}_{h\sim q_{\alpha}}\left[\log (Z_{A_f}(h)/Z_{q,\alpha})\right]+ \mathbb{E}_{h\sim q_{\alpha}}\left[H_{p,\alpha}(h) - H_{q,\alpha}(h)\right]=:\tilde{E}_{q}(\alpha),
\label{eq:feynman_bogoliubov}
\end{equation}
 We can estimate $E(\alpha)$ by taking $q_\alpha$ which minimizes the upper bound (\ref{eq:feynman_bogoliubov}). In the following sections we turn to compute the upper bound. Since we consider only $\alpha\approx 1$, we drop $\alpha$ in what follows. 

When $N_{L-1}$ is large, $\kappa_{A}$  is weakly fluctuating with $h^{L-1}$, since it is a sum of $N_{L-1}$ random variables. Denoting $\overline{\kappa}_{A_{f}}=\mathbb{E}_{h\sim\tilde{p}\left(h\right)}\left[\kappa_{A}\left(h^{L-1}\right)\right]$, we have
\begin{equation}
    \mathbb{E}_{h\sim q_1}\left[H_{p,1}(h)-H_{q,1}(h)\right]\approx\frac{1}{2\overline{\kappa}_{A}}+\mathbb{E}_{h\sim q}\left[H_{\tilde{p}}(h)-H_{q}(h)\right]
\end{equation}
to leading order (where $q \equiv q_1$), so that 
\begin{equation}
     \tilde{E}_{q}\left(1\right)\approx\frac{1}{2\overline{\kappa}_{A}}+\mathbb{E}_{h\sim q}\left[H_{\tilde{p}}\left(h\right)-H_{q}\left(h\right)\right]+\mathbb{E}_{h\sim q_{\alpha}}\left[\log (Z_{A_f}(h)/Z_{q,\alpha})\right].
 \end{equation}
We show in \ref{app:justification_for} that the expectation of log term- $\mathbb{E}_{h\sim q_{\alpha}}\left[\log (Z_{A_f}(h)/Z_{q,\alpha})\right]$ is subleading and thus can be neglected. 
In essence, our variational ansatzes constrain deviations from the GP to a sub-extensive subset of feature directions, such that the resulting corrections do not scale extensively with system size or parameter count. Moreover, since these corrections enter logarithmically, their contribution becomes further subleading.
Thus, we have 
\begin{equation}
    \tilde{E}_{q}\approx\frac{1}{2\overline{\kappa}_{A}}+\mathbb{E}_{h\sim q}\left[H_{\tilde{p}}\left(h\right)-H_{q}\left(h\right)\right]
\end{equation}
The problem has now been reduced to minimizing \(\tilde{E}_q\) with respect to a choice of variational distribution \(q(h)\). If we find the optimal \(q_*\) that minimizes $\tilde{E}_{q}$, we obtain our best estimate for the energy, when $\alpha\approx 1$,
\begin{equation}
    E(\alpha\approx 1) \approx \tilde{E}_{q_*}.
\end{equation}

\subsection{Approximate Variational Energy}
\label{app:layerwise_var}
Our final task is to compute the variational energy $\tilde{E}_q(\alpha)$ given our variational density $q$. Although we work with Gaussian $q$'s that are layer-wise and neuron-wise decoupled, this computation is nevertheless challenging due to the presence of $h^l$, dependence on $[\tilde{K}_{l}]^{-1}$, and the trace term $\Tr \log(\tilde{K}_l)$. In the spirit of kernel adaptation \cite{ringel_applications_2025}, we assume the fluctuations of $\tilde{K}_l$ are sufficiently small and only keep the leading order dependence which dominates $\tilde{E}_q(\alpha)$. In this context, this means replacing the fluctuating kernel $\tilde{K}_{l}$ with its expectation with respect to $q$ and $K_{l}$ given by
\begin{equation}
\label{eq:K_l}
K_{l}\left(x,x'\right)=\frac{\sigma_{l+1}^{2}}{N_{l}}\sum_{i=1}^{N_{l}}\mathbb{E}_{h_i^l \sim q(h^l)]}[\sigma(h_{i}^{l}\left(x\right))\sigma(h_{i}^{l}\left(x'\right))].
\end{equation}
Following this substitution, we obtain
\begin{equation}
H_{\tilde{p}}(h)=\frac{1}{2}\sum_{l=1}^{L-1}\sum_{i=1}^{N_{l}}\langle h_{i}^{l},[K_{l-1}]^{-1}h_{i}^{l}\rangle
\end{equation}
and obtain our final tractable expression for the variational energy
\begin{equation}
\tilde{E}_{q}\propto \frac{1}{2}\left\langle y,K_{L-1},y\right\rangle ^{-1}+\frac{1}{2}\mathbb{E}_{h\sim q}\left[\sum_{l=1}^{L-1}\sum_{i=1}^{N_{l}}\left\langle h_{i}^{l},K_{l-1}^{-1},h_{i}^{l}\right\rangle -H_{q}\left(h\right)\right].
\end{equation}
Since $\tilde{p}$ is fully decoupled layer and neuron-wise, we similarly take a decoupled variational ansatz so that $q(h)=\prod_{l=1}^{L-1}\prod_{i=1}^{N_l}q_{l,i}(h_i^l)$. Further taking $q_{l,i}$ to be Gaussian with mean $\mu_{l,i}$ and variance $Q_{l,i}$, we obtain
\begin{equation}
\label{eq:variational_energy_tractable_gaussian}
  \tilde{E}_{q}=\frac{1}{2}\left\langle y,K_{L-1},y\right\rangle ^{-1}+\frac{1}{2} \sum_{l=1}^{L-1}\sum_{i=1}^{N_{l}} {\left\{\mathbb{E}_{h_{i}^{l}\sim q_{l,i}}\left[\left\langle h_{i}^{l},\left[K_{l-1}\right]^{-1}-Q_{l,i}^{-1},h_{i}^{l}\right\rangle \right]+\left\langle \mu_{l,i},Q_{l,i}^{-1},\mu_{l,i}\right\rangle\right\}}
\end{equation}

\begin{figure}
    \centering
    \vspace{-2.8cm}
    \includegraphics[width=1\linewidth]{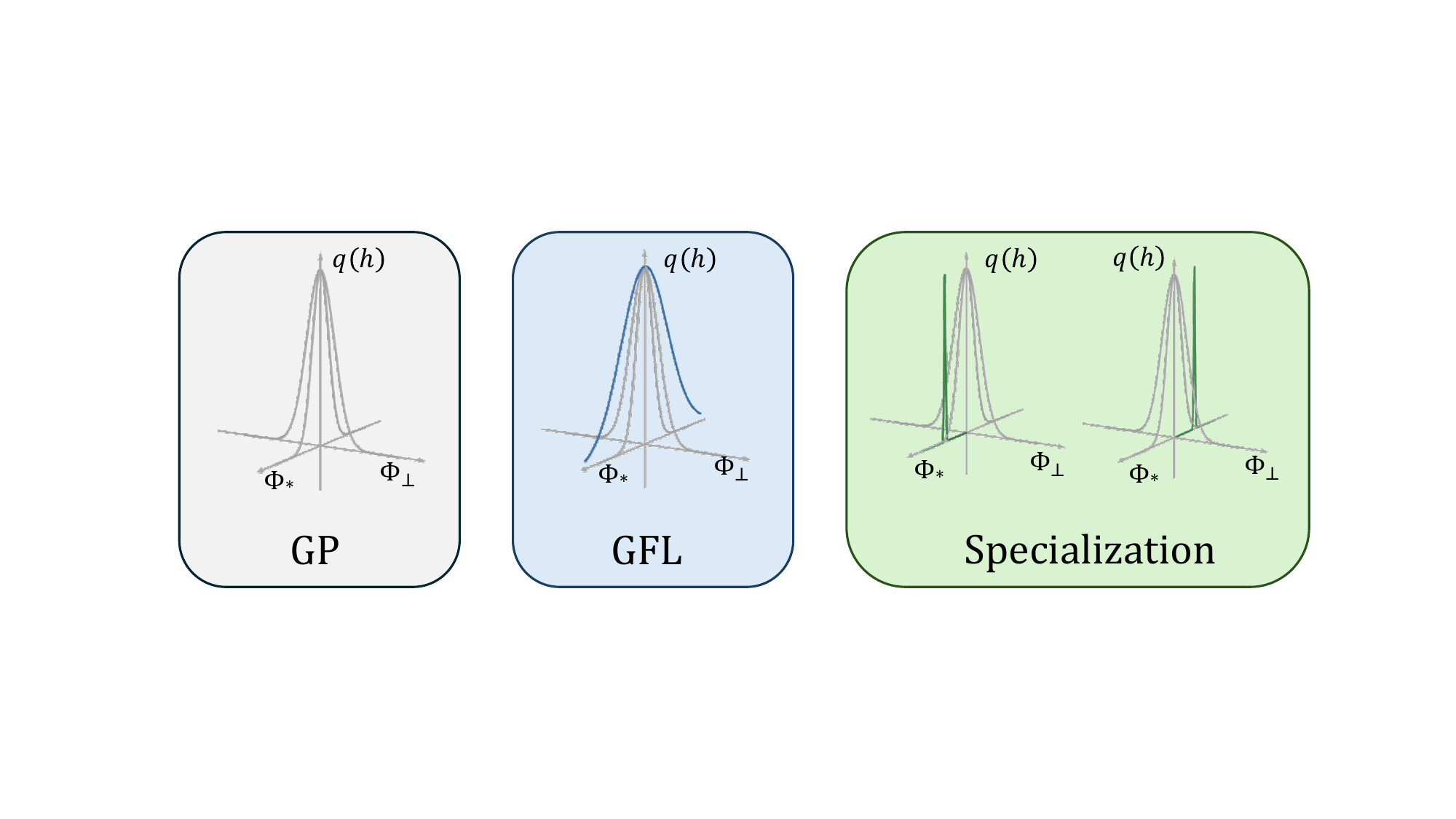}
    \vspace{-2.8cm}
    \caption{Schematic illustration of different candidate feature learning patterns per neuron. }
    \label{fig:fl_schematic} 
\end{figure}

Minimizing the above over all Gaussian $q$'s
results in complex non-linear equations, which we wish to avoid. Instead, we propose minimizing over a small subset of $q$'s, which we refer to as FL patterns. We look to minimize $\tilde{E}_{q}$ over this finite set and take that to be our variational energy. A summary of all patterns can be found in Fig. ~\ref{fig:fl_schematic}.

\subsubsection{Comparison to Adaptive Kernel Approximation }

The similarities and differences between the approach developed here and the adaptive methods discussed in \cite{Seroussi2023,ringel_applications_2025} are worth examining. In those works, $\tilde{p}$ is similarly approximated using a mean-field expansion, with two key differences: The chosen mean and the order of expansion around the mean. The fluctuating kernel in these works $\tilde{K}_{l-1}$ is expanded around its \textit{self-consistent mean} $K_{l-1\mid \tilde{p}}$, rather than the variational expectation introduced here. The self-consistent mean kernel is defined as  
\begin{align}
K_{l>0\mid \tilde{p}}(x,x') &= \frac{\sigma_{l+1}^{2}}{N_{l}} 
\sum_{i=1}^{N_{l}} 
\mathbb{E}_{h_i^l \sim \tilde{p}(h)} 
\!\left[ 
\sigma(h_i^l(x)) \, \sigma(h_i^l(x')) 
\right], \\ \nonumber
K_{0\mid \tilde{p}}(x,x') &= \frac{\sigma_0^2}{d} \, x \cdot x'.
\end{align}
We define the kernel fluctuations relative to this mean by $\Delta K_{l-1}$, such that 
$\tilde{K}_{l-1} = K_{l-1\mid \tilde{p}} + \Delta K_{l-1}$. 
The adaptive approximations assume that these fluctuations are weak, i.e., $\Delta K_{l-1} \ll K_{l-1\mid \tilde{p}}$, 
and retain all terms up to order $\mathcal{O}(\Delta K_{l-1})$, whereas in this work we consider only the zeroth-order term in $\Delta K_{l-1}$. Taking such an approximation in the self-consistent setting would yield the GP distribution, 
since no feature-learning corrections would be taken into account. In our proposed methodology, because the variational expectation already incorporates feature-learning effects, 
the simpler expansion can instead be employed.

\subsection{Justification for Neglecting the Partition Function Ratio}
\label{app:justification_for}
Here we argue that the logarithmic ratio of partition functions, $\mathbb{E}_{h\sim q_{\alpha}}\left[\log (Z_{A_f}(h)/Z_{q})\right]$, is sub-leading relative to the variational energy term $\tilde{E}_q$ (defined in Sec.~\ref{app:variation_est_intro}). We show that this holds for all feature learning patterns considered in this work (Sec.~\ref{sec:feature_learning}): GP, GFL, and specialization. 
Substituting the definition of $Z_{A_f}(h)$ we obtain:
\begin{align}
    \mathbb{E}_{h\sim q_{\alpha}}\left[\log Z_{A_f}(h)\right]&=\mathbb{E}_{q_{\alpha}}\left[\frac{1}{2}\log\kappa_{A}(h^{L-1})\right]+\frac{1}{2}\sum_{l=1}^{L-1}\Tr\mathbb{E}_{q_{\alpha}}\left[\log(\tilde{K}_{l-1})\right] 
\end{align}
We assume that for sufficiently wide networks the fluctuations are weak, so that at least in terms of scales we can approximate
\begin{equation}
    ...\approx \log\mathbb{E}_{q_{\alpha}}\left[\frac{1}{2}\kappa_{A}(h^{L-1})\right]+\frac{1}{2}\sum_{l=1}^{L-1}\Tr\log\mathbb{E}_{q_{\alpha}}\left[(\tilde{K}_{l-1})\right]\nonumber
\end{equation}
So that we obtain
\begin{align}
   \mathbb{E}_{h\sim q_{\alpha}}\left[ \log  Z_{A_f}(h)\right]\approx\frac{1}{2}\sum_{l=1}^{L-1}\Tr \log K_{l-1}+\frac{1}{2}\log{\overline{\kappa}_{A}}
\end{align}

Recalling that $\tilde{E}_q$ scales linearly with $\overline{\kappa}_A^{-1}$, we can indeed justify neglecting the term $\log{\overline{\kappa}_{A}}$ as it is subleading for small $\overline{\kappa}_{A}$. We will show that the remaining terms can be neglected as well. Using the eigendecomposition of the kernel (\ref{eq:K_l}), $K_{l-1}(x,x') = \sum_{k_l} \lambda_{k_l} \Phi_{k_l}(x) \Phi_{k_l}(x')$, the Gaussian integral for a single neuron in layer $l$ evaluates to $\prod_{k_l} \sqrt{2\pi \lambda_{k_l}}$. The logarithm of the full partition function is then
\begin{equation}
    \log Z_p = \sum_{l=1}^{L-1} N_l \log\left(\prod_{k_l} \sqrt{2\pi \lambda_{k_l}}\right) =: \sum_{l=1}^{L-1} N_l \log Z_p^l,
\end{equation}
where we have defined $\log Z_p^l$ as the single-neuron log-partition function under the GP prior. As described in the main text, we consider variational distributions $q$ that fully factorize over layers and neurons. Thus, we can write
\begin{equation}
    \log Z_q = \sum_{l=1}^{L-1} \sum_{i=1}^{N_l} \log \int Dh_{i}^{l} \, q_{l,i}(h_{i}^{l}) =: \sum_{l=1}^{L-1} \sum_{i=1}^{N_l} \log Z_q^{l,i}.
\end{equation}

Combining these results, the log-ratio of the partition functions decomposes into a sum over per-neuron contributions. We aim to show that, for each neuron, this term is sub-dominant compared to the corresponding term in the variational energy $\tilde{E}_q$. Recall (\ref{eq:variational_energy_tractable_gaussian}) that
\begin{equation}
    \tilde{E}_q = \frac{1}{2}\langle y, K_{L}^{-1}, y \rangle + \sum_{l=1}^{L-1} \sum_{i=1}^{N_l} \Delta_{l,i},
\end{equation}
where the per-neuron cost $\Delta_{l,i}$ is 
\begin{equation}
\label{eq:cost_per_neuron}
    \Delta_{l,i} := \frac{1}{2} \mathbb{E}_{h \sim q_{l,i}} \left[  \langle h, K_{l-1}^{-1} - Q_{l,i}^{-1}, h \rangle + \langle \mu_{l,i}, Q_{l,i}^{-1}, \mu_{l,i} \rangle \right].
\end{equation}
Our goal is to show that, for each feature learning pattern, $\Delta_{l,i} + (\log Z_q^{l,i} - \log Z_p^l)$  scales like $\Delta_{l,i}$.

\subsubsection*{Case-by-Case Analysis}

\begin{enumerate}
    \item \textbf{Specialization:} The variational distribution is a delta function
\begin{equation}
    q_{l,i}(\langle h_i^l,\Phi_*^l\rangle)=\delta[\langle h_i^l,\Phi_{*}^l\rangle-\mu_{l,i}],\quad  q_{l,i}(\langle h_i^l,\Phi_{\perp}^l\rangle)=\mathcal{N}(0,\langle\Phi_{\perp}^l ,K_{l-1},\Phi_{\perp}^l\rangle).
\end{equation} 
The parition function is given by- $\log Z_q=\sum_{l=1}^{L-1}\sum_{k_l,\Phi_{k_l}\neq \Phi_*} \log \sqrt{2\pi\lambda_{k_l}}$
    \begin{itemize}
        \item \textbf{Variational Cost:} The cost term evaluates to $\Delta_{l,i} = \frac{1}{2}\mu_{l,i}^2 \langle \Phi_*, K_{l-1}^{-1}, \Phi_* \rangle$. In typical high-dimensional settings, $\langle \Phi_*, K_{l-1}^{-1}, \Phi_* \rangle$ scales at least with $d$, so $\Delta_{l,i} = \mathcal{O}(d)$.
        \item \textbf{Partition Function Ratio:} In all but the feature direction, the normalization factor of the variational estimate and the normalization factor of $p$ is equal, and thus cancels out. The only difference is in the feature direction, where we obtain exactly $\log Z_p^l/Z_q^l = -\frac{1}{2} \sum_{i=1}^{N_l}\log(\Delta_{li})$ 
        \item \textbf{Conclusion:} we thus obtain $Z_p^l/Z_q^l \ll \sum_{i=1}^{N_l}\Delta_{il}$.
    \end{itemize}

    \item \textbf{Gaussian Feature Learning (GFL):} We have $h \sim \mathcal{N}(0, Q_{l,i})$, where $Q_{l,i}$ shares the same eigenbasis as $K_{l-1}$, with $\lambda'_{k_l}$ identical to $\lambda_{k_l}$ except for the feature $\Phi_*$, where $\lambda'_* = D \lambda_*$.
    \begin{itemize}
        \item \textbf{Variational Cost:} The cost is $\Delta_{l,i} = \frac{1}{2}\text{Tr}(K_{l-1}^{-1}Q_{l,i} - I)$. Since only one eigenvalue differs, this sum reduces to $\Delta_{l,i} = \frac{1}{2}(D-1)$.
        \item \textbf{Partition Function Ratio:} The log-ratio $\log(Z_q^{l,i}/Z_p^l)$ becomes the log-ratio of the determinants, which is $\frac{1}{2}\log(\det Q_{l,i} / \det K_{l-1}) = \frac{1}{2}\log(D)$.
        \item \textbf{Conclusion:} For large enhancement factors $D$, the linear scaling of $\Delta_{l,i} = \mathcal{O}(D)$ dominates the logarithmic scaling of the partition function term, $\mathcal{O}(\log D)$.
    \end{itemize}

    \item \textbf{GP / Lazy Regime:} This case is equivalent to GFL with $D=1$.
    \begin{itemize}
        \item \textbf{Conclusion:} Substituting $D=1$ into the GFL results, we find that both $\Delta_{l,i} = 0$ and the log-partition function term is $\log(1) = 0$. The relationship holds trivially.
    \end{itemize}
\end{enumerate}

\subsection{Kernel Feature Propagation}
\label{app:feature_prop}
Since the cost of each layer depends on the kernel of the previous layer, an important element of our heuristic is understanding how the choice of pattern in the previous layer affects the kernel and its spectrum. Recall that we denote the pre-activation of the current layer by $h^l(x)$, and of the previous layer by ${h}^{l-1}(x)$,  where ${h}^{l-1}(x)$ is the width $N_{l-1}$ pre-activation vector of the previous layer. Let $h^{l-1}_i$ be distributed according to some candidate Gaussian distribution $q^{\prime}_i(h^{l-1})$. We wish to understand how this choice of $q^{\prime}_i(h^{l-1})$ affects the kernel of the next layer given by  
\begin{align}
K_{l-1}(x,x') &= N_{l-1}^{-1} \sum_{i=1}^{N_{l-1}} \mathbb{E}_{h^{l-1}_i(x) \sim q_{i} }[\sigma(h^{l-1}_i(x))\sigma(h^{l-1}_i(x'))]=:\sum_{i=1}^{N_{l-1}}{K_{l-1,i}(x,x')},
\end{align}
where we took $\sigma_l^2=1$ for brevity.

\textbf{Claim ({\romannumeral 1}): Neuron specialization creates a spectral spike.} 
Assume we have an operator of the form $K(x,x')=A(x,x') + c\sigma(\Phi(x))\sigma(\Phi(x'))$ for some constant $c$. We wish to understand the behavior of the RKHS norm of $\sigma(\Phi)$ with respect to the above operator. Using the Sherman-Morrison formula, we obtain
\begin{align}
[A + c\sigma(\Phi)\sigma(\Phi)^\top]^{-1} &= A^{-1} - \frac{A^{-1} c\sigma(\Phi)\sigma(\Phi)^\top A^{-1}}{1+ c \sigma(\Phi)^\top A^{-1} \sigma(\Phi)},
\end{align}
where $vv^\top$ denotes the outer product. Consequently, the RKHS norm of $\sigma(\Phi)$ with respect to $K$ is given by
\begin{align}
R_{K}\equiv \sigma(\Phi)^\top [A + c\sigma(\Phi)\sigma(\Phi)^\top]^{-1} \sigma(\Phi) = R_A - \frac{R_A^2c}{1+R_Ac} = \frac{R_A}{1+R_Ac} < 1/c
\end{align}
where $R_A$ is its RKHS norm w.r.t. $A^{-1}$. Typically, we would consider large values of $R_A$. Thus, if we have $c R_A \gg 1$, then- $\frac{R_A}{1+R_Ac}\approx c^{-1}$, thereby justifying Claim (i).

\paragraph{Corollary 1.}
If $M$ of $N_l$ neurons are specialized in a given layer with the rest remaining Gaussian with zero mean, then we can substitute $c=N_l/M$ in the above, and obtain $R_{K_{l}}=N_l/M$.

\paragraph{Corollary 2.}
If all neurons are specialized with proportionality constant $\sqrt{\beta}$, then we can substitute $c=\beta$ in the above, and take $A= \epsilon I$ as a regularizer, with $\epsilon \rightarrow 0$, so that $R_A\rightarrow\infty $, and we obtain $R_{K_{l}}=\beta$.

\textbf{Claim ({\romannumeral 2}): Amplified features in the pre-activation kernel create amplified higher-order features in the post-activation kernel.}
We begin by considering the Gaussian case, where $h^{l}_i(x)\sim \mathcal{N}(0,Q_{l})$ for all $i$. Since $K_{l}$ is a dot product kernel for an FCN, we can expand the expectation in (\ref{eq:K_l}) to obtain 
\begin{align}
K_{l}(x,x') 
&=\sum_{a=1}^{\infty} (Q_{l}(x,x'))^a\  F_a(x,x'),
\end{align}
where $Q_{l}(x,x')$ is a function of $x \cdot x'$ and $F_a(x,x')$ is a function of $|x|^2$ and $|x'|^2$
derived by expanding the resulting Cho and Saul kernel associated with the activation function $\sigma$ \cite[\S 2.3]{Cho}. We further assume that $Q_{l}(x,x)\sim\mathcal{O}(1)+$small fluctuations. This is a reasonable assumption to make for Gaussian data in high dimensions where $|x|^2 = d \pm \sqrt{d}$. For more general datasets, one should either verify this or apply layer-normalization (as done below for the case of data with power-law covariance spectrum). 

Using its Mercer decomposition $Q_{l}(x,x')=\sum_{k} \lambda_k \Phi_{k}(x) \Phi_{k}(x')$ , we have 
\begin{align}
K_{l}(x,x') &\propto \sum_{k} \lambda_{k} \Phi_{k}(x) \Phi_{k}(x') + \sum_{k_1,k_2} \lambda_{k_1} \lambda_{k_2} \Phi_{k_1}(x) \Phi_{k_2}(x) \Phi_{k_1}(x') \Phi_{k_2}(x')+... 
\end{align}
Next, we ask how enhancing a certain $\lambda_k$ by a factor $D$ affects the post-activation kernel $K_l(x,x')$. Let $\lambda_{k_*}$ be some eigenvalue of $Q_{l}$ corresponding to the eigenfunction $\Phi_{*}(x)$. It can be potentially adapted via GFL ($\lambda_{k_*} \rightarrow D \lambda_*$ where $D \lambda_* < 1$). Then the $K_{l,i}$ RKHS norm of  any feature of the form $\sum_{n=1}^m a_n \Phi^n_{\lambda_*}(x)$ scales like $\mathcal{O}((\lambda_* D)^{-m})$ (assuming $a_m=\mathcal{O}(1)$). Indeed, consider the  case $m=2$ and examine
\begin{align}
 K_{l}(x,x')  \propto  \sum_{k} D_k\lambda_k \Phi_{k}(x) \Phi_{k}(x')   +  \sum_{k_1,k_2} D_{k_1}D_{k_2} \lambda_{k_1}  \lambda_{k_2} \Phi_{k_1}(x) \Phi_{k_2}(x) \Phi_{k_1}(x') \Phi_{k_2}(x')  + ...
\end{align}
with $D_{k}=D$ when $\lambda_k = \lambda_*$ and $D_{k}=1$ otherwise. The argument proceeds by treating the term $\Phi^m_*(x)\Phi^m_*(x')$ as a rank-1 spike. 
Substituting $K_l$  for $K$, $(D\lambda_*)^m$ for $c$, and $\Phi^m$ for $\sigma(\Phi)$ in Claim (i), we obtain $R_{K_l}=(R_A^{-1}+(D\lambda_*)^{m})^{-1}$. 

\begin{figure}[h!]
    \includegraphics[width=1\linewidth]{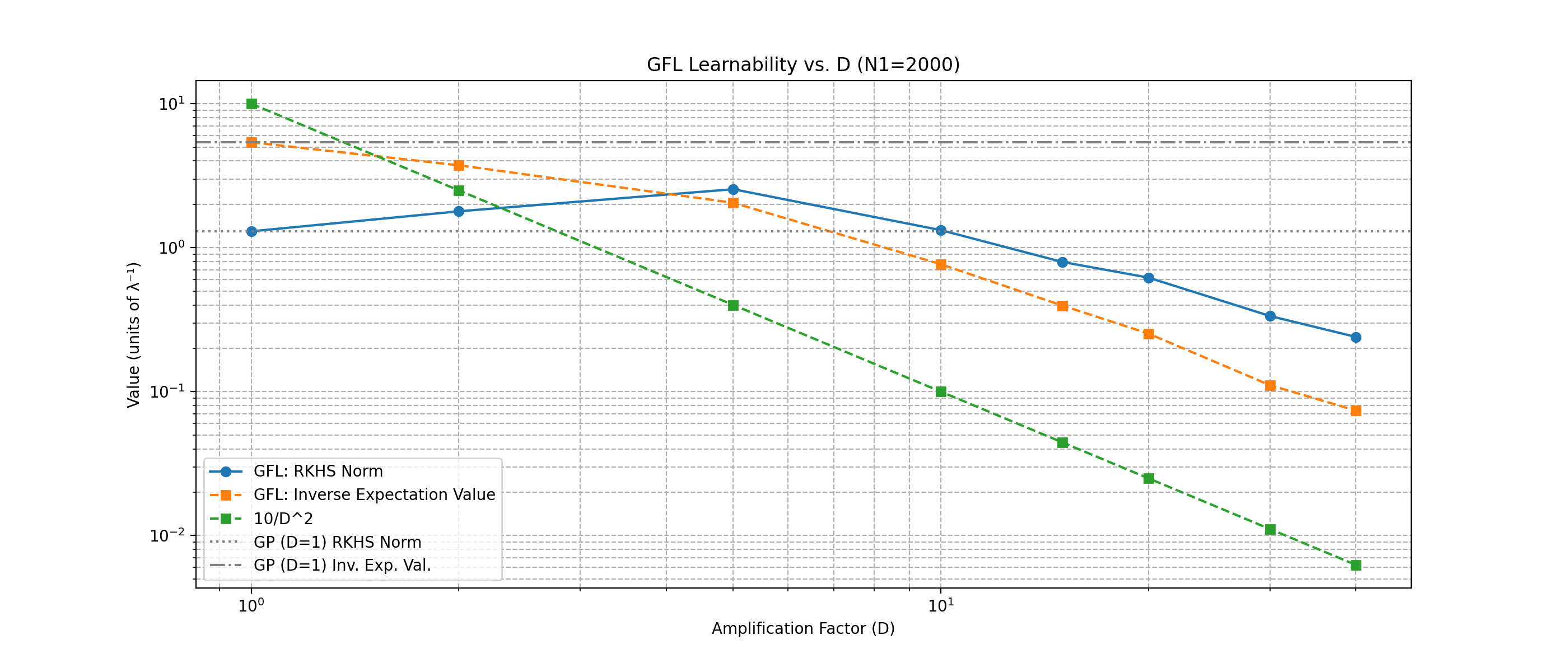}
    \caption{Demonstration GFL feature propagation. We increased the variance of the $d/2$ highest kernel mode $\Phi_*(x)$ of the first hidden layer and then measured $\langle \Phi_* | K_{l=3} | \Phi_* \rangle^{-1}$ and $\langle \Phi_* | [K_{l=3}]^{-1} | \Phi_* \rangle$, where $K_{l=3}$ is the kernel of the subsequent layer. We used ReLU activations, $d=120,N_1=2000,N_2=1000$ and random Gaussian data. This demonstrates the expected $D^{-2}$ decay and the matching between inverse expectation values and the RKHS. 
    \label{fig:FeaturePropagation_GFL}
    } 
\end{figure}

\begin{figure}[h!]
\includegraphics[trim={0cm 0 9cm 0},clip,width=1\linewidth]{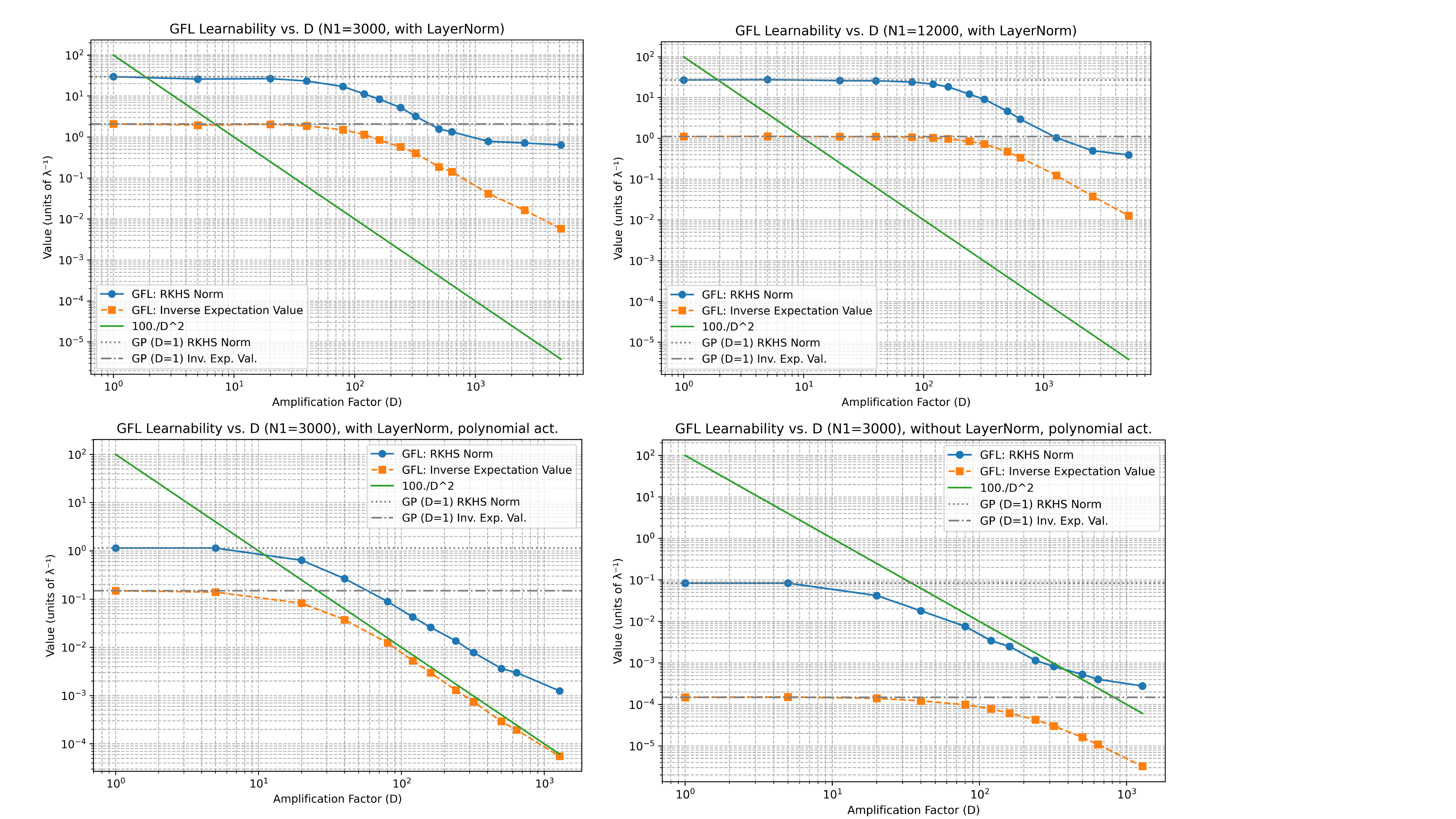}
    \caption{Similar to Fig.~\ref{fig:FeaturePropagation_GFL}, only for Gaussian data with power-law covariance matrix with eigenvalues decaying as $k^{-1.1}$. See \ref{app:feature_prop} for further details.}
\label{fig:FeaturePropagationGFLpowerlaw}
\end{figure}

The important observation/assumption here is that, for sufficiently large $D$, $R_A (D \lambda_*)^m \gg 1$ so that $R_{K_l}\approx (D\lambda_*)^{-m}$. This is supported numerically in Figs. \ref{fig:FeaturePropagation_GFL} and \ref{fig:FeaturePropagationGFLpowerlaw}. Our analytical reasoning is that the subspace effectively described by the kernel $K_l$ is a negligible portion of the total function space. This is either because the rank of $K_l$ is constrained by network width $N_l$ or its spectrum $\lambda_k$ exhibits rapid decay.  Our analytical rationalization for this is as follows. Either because in a real network, the rank of the kernel $K_l$ is bounded by the network width $N_l$ or because the spectrum $\lambda_k$ falls down quickly, the effective span of the first term and subsequent terms covers only a negligible fraction of the total function space. 
In this large function space, it is unlikely that a feature like $\Phi^m_*$ is fully contained within this span. If this is the case, its RKHS norm with respect to the base kernel would be large, motivating our observation/assumption. 

In Fig.~\ref{fig:FeaturePropagation_GFL}, we consider two ReLU-activated layers of an FCN ($L=4$) and Gaussian i.i.d. data $x \in \R^d, x \sim {\gN}(0,I_d)$. We obtain the empirical NNGP kernel after the first ReLU layer of width $N_1=2000$. Diagonalizing the kernel ($K_{l=2}$) on $P=6000$ sample points $\{x_{\nu}\}_{\nu=1}^P$, we choose $\Phi_*(x)$ to be the $d/2$ highest eigenvalue. We then look for the best weight vector $a \in \R^{N_1}$ satisfying $\Phi_*(x_{\nu}) \approx \sum_{i=1}^{N_1} a_i \mathrm{ReLU}(W_i \cdot x_\nu)$, where $W_i$ are the input layer weights used to generate $K_{l=2}$. We draw the rows of $W \in \R^{N_2 \times N_1}$ i.i.d. with $W_i \sim \gN\left(0,I/N_1 + (D-1)\hat{a} \hat{a}^\top/N_1\right)$, where $\hat{a}=a/\sqrt{a^\top a}$. This mimics the GFL effect. Finally, we compute the kernel empirical kernel following another ReLU layer and compute the RKHS norm of $\Phi_*^2(x)$ with mean removed, omitting eigenvalues which are zero up to machine precision (for $P > N_2$, $P-N_2$ such eigenvalues are to be expected).  

In Fig.~\ref{fig:FeaturePropagationGFLpowerlaw}, we perform a similar analysis with the following qualitative changes. The data is chosen to be Gaussian, but non-i.i.d. with a covariance matrix whose $k$'th eigenvalue is $k^{-1.1}$. To prevent strong fluctuations of $K_2(x,x)$ (i.e., $Q_2(x,x)$ of Claim (ii)), which are now not suppressed at $d \rightarrow \infty$ we apply layer normalization between the two activated ReLU layers. The top left panel has $d=250,N_1=N_2=3000$ and uses $4000$ datapoints to sample the kernel operator and compute the spectrum used in the RKHS norm. The top right panel is a $\times 4$ scaled-up version of the left panel in terms of $N_1,N_2,P$ and $d$ and shows a longer persistence of $1/D^2$ expected scaling by, roughly, a factor of $2$. The bottom panels show the behavior of polynomial activation of the type $\sigma(x)=1+x+x^2$, both with (left) and without (right) layer normalization. 

\section{\texorpdfstring{$\tilde{E}$ and $\Delta_\ell$ in Weight Space}{E-tilde and Delta-ell in weight space}}
\label{App:WeightSpace}

In the main text, we focused on a function space formulation for $\tilde{E}$, using pre-activations, GP distributions ($q$) for $h^l$, and kernel operators. This formulation facilitated the combination of GP-like learning mechanisms and circuit/specialization-based mechanisms. In some cases however, it is more natural to work in weight space (see also \cite{Guth2024} for a weight-space picture analogous to kernel adaptation). One such example is CNNs, for which the distribution of the input layer convolutional-patch-dependent pre-activations, is more readily described in terms of shared input weights. Another example would be an $i_0$-th neuron specializing on some $j_0$ neuron of the upstream layer, which implies a delta-function distribution around $W_{i_0 j} \sim \delta_{j,j0}$. Accordingly, we wish to express $\Delta_{l,i}$ of (\ref{eq:cost_per_neuron}) in weight space. For simplicity, we focus on distributions with zero mean. 

To this end, we consider some pre-activation in layer $l$ and write it in terms of layer weights namely, $h^l_{i_0}(x) = \sum_{i=1}^{N_{l-1}} W^l_{i_0 i} \sigma(h^{l-1}_i(x))$. For compactness, we henceforth take $W^l_{i_0 i} \rightarrow a_i,h^l_{i_0} \rightarrow h^l$. We next treat the $h^{l-1}_i$'s as independent draws from $q_{l-1,i}=q_{l-1}$, consistent with the kernel adaptation approximation, which removes correlations between pre-activation.
To avoid operator algebra and work solely with matrices, we consider extremely many ($P'$) draws $\{x_\nu\}_{\nu=1}^{P'}$
from $d\mu_x$ and define the ($l-1$)-th feature matrix with entries
\begin{align*}
    F_{i \nu}^{l-1} := \sigma(h^{l-1}_i(x_{\nu})), \quad i=1,\ldots,N_{l-1}, \quad \nu=1,\ldots,P'
\end{align*}
Note that for large $N_{l-1}$, $\left[\left(F^{l-1}\right)^\top F^{l-1}\right]_{\mu \nu}/N_{l-1}$, via the law of large numbers, concentrates to its averages under $q$ and thus approaches $K_{l-1}(x_{\mu},x_{\nu})$.

We next revisit the inter-layer action for $l,l-1$ appearing Eq. (\ref{Eq:InterLayerActions} given by $\langle h, \tilde{K}^{-1},h\rangle$ and note that (i) $\tilde{K}$ upon freezing $h^{l-1}$ at some typical value as done above, $\tilde{K}= F^\top F/N_{l-1}$ (ii) the inverse operation is to be understood as a pseudo inverse ($(..)^{+}$) which removes the contribution of the null space of this (generically) rank-$N_{l-1}$ operator. Following this and the linear relation between $h^l$ and $a$, we have that  
\begin{align}
\langle h,\tilde{K}_{l-1}^{+},h\rangle &= a^T F [F^\top F/N_{l-1}]^+ F^\top a = N_{l-1} a^\top a.
\end{align}
We thus obtain a reasonable result: The RKHS term, which regulates the pre-activations, is equal to the weight-decay term associated with the $a$'s that generated that $h$. 

Following this rewriting, one can now repeat our variational approximation in weight space.This results in an analogous formula to Eq. \ref{eq:est_E} with $q_l$ denoting a Gaussian distribution on $a$'s ($a \sim {\cal N}[\bar{a}^l,\Sigma_l]$), and $K_{l-1}$ (the covariance of the $h$'s in the action) replaced by $I/N_{l-1}$ (the covariance of $a$'s in the action).

All in all we obtain the following weight space version of the contribution of the $i$'th neuron to $\tilde{E}_q$, namely 
\begin{align}
\Delta_{l,i} &= \mathbb{E}_{a \sim \mathcal{N}(\bar{a},\Sigma_{l})} \left[a^\top [N_{l-1}I-\Sigma_l^{-1}] a\right]+ \bar{a}^T \Sigma_l^{-1} \bar{a} 
\end{align}


\section{Application of Heuristics for Additional Architectures}

\subsection{Regression tasks in Two-layer Networks}
\label{sec:two_layer_network}
In the two-layer setting, an exact solution can be obtained, so we begin by comparing our heuristic approach to the exact solution. In this case, we consider both two-layer FCNs as well as CNNs with non-overlapping convolution windows. Together, these are given by
\begin{equation}
    f(x) = \sum_{i=1}^{N_w} \sum_{j=1}^{N} w^{(2)}_{ij} \, \sigma(w^{(1)}_j \cdot x_i),
    \label{eq:app_network}
\end{equation}
where $x \in \mathbb{R}^d$ is drawn from $\mathcal{N}(0,I_d)$. We take $d = N_w S$ so that $w_j, x_i \in \mathbb{R}^S$. The vector $x_i$ is given by the $((i-1)S+1)$-th to $iS$-th coordinates of $x$. We train these networks on a polynomial target of degree $m$ given by $y(x)=\sum_{i=1}^{N_w} He_m(w_* \cdot x_i)$ where $He_m$ is the $m$-th probabilist  Hermite polynomial, which is the standard polynomial choice under our choice of data measure, and $w_*\in \mathbb{R}^S$  is some normalized vector. The networks are trained via Lengevin dynamics \cite{welling_bayesian_nodate}, with ridge parameter $\kappa$,  quadratic weight decay,  and standard scaling. For an extension to mean-field scaling, see App. \ref{app:mf_scaling}.  

\subsubsection{Fully Connected Networks}
\paragraph{Standard scaling.}
We turn to compute $\tilde{E}_q$ for a range of feature learning patterns. For this shallow FCN (and $N_w=1$), our choice of feature learning patterns amounts to considering distributions in a single layer. We consider the following three scenarios (though more combinations are possible): (1) all neurons are GP distributed, (2) all are GFL distributed with amplification $D$, and (3) $M$ neurons specialize on a the same feature, while those remaining are GP distributed. As the kernel of the first layer can only express linear features, the only relevant feature to be considered for the GFL and M-specialization patterns is $\Phi_*(x) = w_* \cdot x$.  We compute the scale of the optimal variational energy for each pattern:

\begin{enumerate}[leftmargin=*,label=(\arabic*)]
    \item {\bf GP}: Here, $\Delta_{1,i}=0$ since $K_{l-1} = Q_{l}$. In this baseline setting, learning $m>1$ is hard since $\langle He_m | K_2 | He_m \rangle = \mathcal{O}(d^{-m})$ (see Sec.~\ref{sec:kernel_prop}). Thus,  in total, we have $\tilde{E}_{q\sim\text{GP}}\propto d^m$.
    \item {\bf GFL}: Following (\ref{eq:est_E}), this pattern incurs a cost of $\Delta_{1,i} =D$ per neuron $i$, resulting in a total cost of $ND$.  The $a_y$ term can be calculated utilizing Claim ({\romannumeral 2}). This leads to a $D^m$-factor decrease in the RKHS norm relative to the GP, so that $a_y\propto (d/D)^m$.
    In total, we find that $\tilde{E}_{q\sim \text{GFL}}\propto ND+(d/D)^m$. Minimizing w.r.t. $D$, we obtain  $D_{\min} = (d^m/N)^{1/(m+1)}$, and, substituting back, we obtain $\tilde{E}_{q\sim \text{GFL}}\propto (Nd)^{\frac{m}{m+1}}$.
    \item {\bf M-Specialization}: Following \ref{eq:est_E}, this pattern incurs a cost of $\Delta_{1}=M\langle \Phi_*,K_0^{-1} \Phi_*\rangle=Md$, where we denote $\Delta_{l}=\sum_i \Delta_{l,i}$.   Utilizing Claim ({\romannumeral 1}), this results in adding a spike with an $M/N$ coefficient along $\sigma(w_* \cdot x)$ in $K_1$  appearing in $a_y$. Before this spike, $He_m(x)$ only had overlaps with the $m$-th order Taylor expansion of the kernel, leading to a $d^{-m}$ scaling. However, since $\sigma(w_* \cdot x)$ has an $\mathcal{O}(1)$ overlap with $He_m(x)$, so that $a_y \propto N/M$, we obtain $\tilde{E}_{q\sim\text{M-Sp}} \propto dM+N/M$. Minimizing further over $M$, the number of specializing neurons leads to $M_{\min}=\sqrt{N/d}$ and therefore $\tilde{E}_{q\sim\text{M-Sp}} \propto \sqrt{dN}$. 

\end{enumerate}

Now, we can compare the different feature learning patterns. Taking the most common linear scaling where $N \propto d$, the specialization scenario has the lowest variational energy. Our scaling theory then predicts an $\mathcal{O}(d)$ sample complexity as well as multimodal distribution of $w$ along $w_*$ with $\mathcal{O}(1)$ specializing neurons. Taking $m > 1$ and $N \gg d^5$, lazy learning wins and leads to $\mathcal{O}(d^m)$ complexity. When $m=1$, GFL and M-specialization are on par for $N \propto d$.  
These calculations coincide with both experimental and direct LDT results, as demonstrated in Fig.~\ref{fig:2_layer_exp} for networks trained on $He_3$. In terms of sample complexity, both predictions agree with experiment, with a scaling of $P_*\propto d$, as seen in Fig.~\ref{fig:2_layer_exp} (b). Our heuristic approach correctly predicts the scaling of the number of specializing neurons with $N$, as seen in Fig.~\ref{fig:2_layer_exp} (c). Finally, as shown in panel (a), the analytical LDT method recovers the correct pre-activation distribution, which corresponds to $q(h)$ for $q\sim\text{M-Sp}$.

\paragraph{Extension to Mean-field Scaling}
\label{app:mf_scaling}
Here we extend the results presented in the main text to the case of mean-field scaling.
We note that the results presented for standard scaling can change when introducing mean-field parametrization, amounting here to enhancing the alignment factor by a $\chi=\mathcal{O}(N)$ factor. The GP pattern then results in a much worse energy $\tilde{E}_{GP}=N d^3$. Revisiting the GFL pattern, we now find $\tilde{E}_{GFL}=ND+N (d/D)^m$ leading to $D \propto d^{1/(m+1)}$ and $\tilde{E}_{GFL,optimal} \propto N d^{m/(m+1)}$. For specialization, we obtain $\tilde{E}_{sp}=dM+N^2/M$, leading to $M \propto N/\sqrt{d}$ and $\tilde{E}_{sp,optimal} \propto \sqrt{d} N$. Taking $N \propto d$, the specialization scenario again wins over for $m>1$, with a sub-extensive ($\mathcal{O}(\sqrt{d})$) number of specializing neurons. Unlike with standard-parametrization, we see that even when taking $N\rightarrow \infty$, we remain in the rich regime. However, at least in this Bayesian finite-ridge setting, sample complexity is better with standard parametrization.

\subsubsection{CNNs with Non-overlapping Patches.}
\label{Sec:CNN}
Our approach can be extended to CNNs as defined by the following equation
\begin{equation}
    f(x) = \sum_{i=1}^{N_w} \sum_{j=1}^{N} w^{(2)}_{ij} \, \sigma(w^{(1)}_j \cdot x_i),
\end{equation}
with $N_w > 1$. In this case, it is better to focus on the covariance of $w_i$, namely, $\Sigma = N^{-1} \sum_{i=1}^{N} w_i w^T_i$, than on the covariance of pre-activations on each path. One can then show that the cost becomes $\Delta_{1} = {\rm E}_{q_w} w^\top \left[\Sigma^{-1}-I_S/S\right] w$. 
\begin{enumerate}[leftmargin=*,label=(\arabic*)]
    \item {\bf GP}: In this scenario, the output kernel is given by $K_{2,\text{CNN}}(x,x') = N_w^{-1} \sum_{i=1}^{N_w} K_{2,\text{FCN}}(x_i,x_i')$, where $K_{2,\text{FCN}}$ is the FCN kernel ($N_w=1$). Focusing on a linear target for simplicity, we can work out the scaling of the relevant (linear) kernel feature by Taylor expanding $K_{2,\text{FCN}}= a_1 x_i \cdot x'_i/S$ to get $K_{2,\text{CNN}} = \frac{a_1}{N_w S} x \cdot x'$, with $a_1$ being some $\mathcal{O}(1)$ constant. Lazy learning then yields $\tilde{E}_{q\sim\text{GP}}=N_w S = d$, as in a FCN with no weight sharing.  
    \item {\bf GFL}: Here, we take $\Sigma = I_S/S + D w_* w_*^\top$, resulting in $\Delta_{l=1} = N D$. The leading term of the Taylor expansion now equals $K_{2,\text{CNN}}= \frac{1}{N_w S} \sum_{i=1}^{N_w} x_i^\top \Sigma x'_i$ leading to a $D/(N_w S)$ scaling of the target. All in all, we find that $\tilde{E}_{q\sim\text{GFL}}=N D + (N_w S)/D$, leading to $\tilde{E}_{q_*}=\sqrt{N M_w S}$ for the optimal $q_*\sim\text{GFL}$. 
    \item {\bf M-Specialization}: In this case, we obtain a $\Delta_{l=1} = M S$ cost. The contribution of the $M$ specializing neurons to the kernel goes as $\frac{a_1 M}{N_w N} \sum_{i=1} (w_* \cdot x) (w_* \cdot x')$, leading to $a_y = \langle y, K_L, y \rangle=\mathcal{O}(M/N_w N)$. Thus, $\tilde{E}_{q\sim\text{M-Sp}}=M S + N_w N /M$ resulting in the optimal variational energy $\sqrt{S N_w N}$.
\end{enumerate}

Both GFL and M-Specialization patterns, in the proportionate limit $N \propto N_w \propto S\rightarrow\infty$, lead to $P_* \propto S^{3/2} = d^{3/4}$. This recovers results reported in \cite{ringel_applications_2025} computed via a mean-field approach.

\subsection{Classification Tasks}
\label{app:classification_relu_parity}
We consider a fully connected network as defined in ~\ref{eq:app_network} with $N_w=1$ and ReLU activations, trained on a $k$-parity classification task. Explicitly, this task is defined as- \begin{equation}
    y_k(x)=\text{sign}\prod_{j=1}^kx_j,
\end{equation} 
where $\eta\sim\mathcal{N}(0,1)$, and for $i=1,..,d$,  $x_i=s_i+\mathcal{N}(0,\epsilon)$, for and $s_i$ that are i.i.d. random binary valuables in $\{\pm1\}$. We take cross-entropy loss, and consider $k=2$. 

Here we observe that the target is given by- 
\begin{equation}
    y_{2}(x)=\text{sign}((x_1+\eta_1)(x_2+\eta_2))=\text{sign}(\frac{1}{2}(x_1+x_2)^2-x_1^2-x_2^2)
\end{equation}
This setting is similar to the previously discussed polynomial target in the erf networks. The difference in this case is that there are multiple relevant directions, namely:  $w_{*,1}=\hat{e}_1,w_{*,2}=\hat{e}_2,w_{*,1+2}=\hat{e}_1+\hat{e}_2$, and up to multiplicative constants on the target and constant additions.  As in the previous cases, we compute the scale of the optimal variational energy for each pattern, and consider the same patterns as in the two layer network. In this case, we can further consider another setting of feature learning. Rather than taking $M$ specializing neurons, and setting each of them to specialize with amplitude $\mu=1$, we can set $\approx1$ neuron to specialize with magnitude $\mu$. As the erf was a bounded activation, this choice was not beneficial in that case we did not need to consider it as well. Defining the last pattern as $\mu$-Specialization, and comparing to the previously discussed patterns, we obtain:

\begin{enumerate}[leftmargin=*,label=(\arabic*)]
    \item {\bf GP}: Here, $\Delta_{1,i}=0$ since $K_{l-1} = Q_{l}$. As in the case of the erf network, we have $\tilde{E}_{q\sim\text{GP}}\propto d^2$.
    \item {\bf GFL}: Here we have $\tilde{E}_{q\sim \text{GFL}}\propto ND+(d/D)^2$. Minimizing w.r.t. $D$, we obtain  $D_{\min} =d^{2/3}/N^{1/3}$, and, substituting back, we obtain $\tilde{E}_{q\sim \text{GFL}}\propto (Nd)^{\frac{2}{3}}$.
    \item {\bf M-Specialization}: $\tilde{E}_{q\sim\text{M-Sp}} \propto dM+N/M$. Minimizing further over $M$, the number of specializing neurons leads to $M_{\min}=\sqrt{N/d}$ and therefore $\tilde{E}_{q\sim\text{M-Sp}} \propto \sqrt{dN}$. 

    \item {\bf $\mu$-Specialization}: Following \ref{eq:est_E}, and the second feature propagation rule, we obtain this pattern incurs a cost of $\Delta_{1}=\mu^2d$.  Utilizing the first feature propagation rule, this pattern results in adding a spike of the kernel so that $a_y \propto N/{\mu^2}$, we obtain $\tilde{E}_{q\sim\text{M-Sp}} \propto d\mu^2+N/\mu^2$. Minimizing further over $\mu$, the magnitude of the specializing neurons leads to $\mu_{\min}= (N/d)^{1/4}$ and therefore $\tilde{E}_{q\sim\text{$\mu$-Sp}} \propto \sqrt{dN}$. 

\end{enumerate}

\begin{figure}
    \centering
    \includegraphics[width=1.\linewidth]{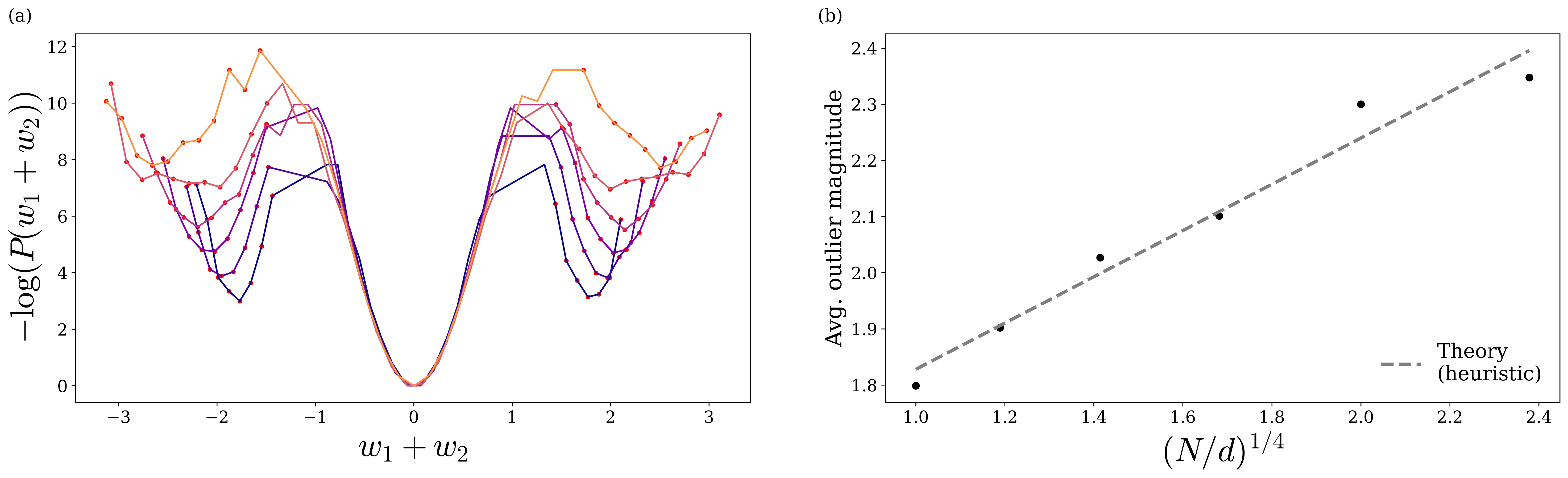}
    \caption{Emergence of specialization in two layer ReLU networks trained on classification task. In (a), the negative log distribution of the weights in the hidden layer in one of the feature directions is shown, with neurons classified as specializing appearing in red. Specifically, in this figure the distribution is shown for $w_{*,1+2}$. Although in this  setting there are multiple relevant feature direction, we observe a similar $\mu$-Specialization behavior in all of these. A single direction is considered here for simplicity. In (b) the magnitude of the specializing neuron as a function of the network width is shown. Indeed, our Heuristics manage to correctly predict the scaling of the learned feature with increasing here. }
    \label{fig:relu_fig}
\end{figure}

There is a redundancy in the preferential patterns, as both specialization distributions result in a variational energy that scales as $P_*\propto \sqrt{Nd}$.  Experimentally, we find that indeed the feature learning patterns that emerge for this case correspond to $\mu$-Specialization, with the correct scaling of the specialization magnitude with the layer width, as can be seen in ~\ref{fig:relu_fig}.

 \subsection{Softmax Attention Layer}
\label{App:SoftMaxLayer}
Here, we consider an attention block of the form 
\begin{align}
f(X) &= \frac{1}{\sqrt{L}} \sum_{h=1}^H \sum_{a,b=1}^L @_{ab;h}(X) (w_h \cdot x^b) \\ \nonumber 
@_{ab;h}(X)&= \frac{ e^{[x^a]^\top A_h x^b}}{\sum_{c=1}^L e^{[x^a]^\top A_h x^c}}
\end{align}
where $X \in \R^{L \times d}, A_h \in \R^{d \times d}$, $x^a \in \R^d$ is the $a$-th row of $X$, and $w_h \in \R^d$. Our prior over network weights is $\prod_{h=1}^H {\cal N}[0,I_{d^2}/d^2; A_h] {\cal N}[0,I_{d}/(dH);w_h]$. The only context length ($L$) dependence comes from the pre-factor of $1/\sqrt{L}$ which ensures that when $X^a_i \sim \gN(0,1)$ then we have $f(X) = \mathcal{O}(1)$. We consider the target function $y(X) = \sum_{a,b} \frac{1}{\sqrt{L (L-1)}}x^a_1 x^b_2 x^b_3$. 

As $H \rightarrow \infty$, the above $f(X)$ tends to a GP with $w_h$ taking the role of read-out layer weights and a kernel given by
\begin{align}
K(X,X') &= (dL)^{-1} \sum_{a,a',b,b'=1}^L (x^b \cdot (x')^{b'}) \mathbb{E}_{A} \left[  @_{ab}(X) @_{a'b'}(X')\right],
\end{align}
where we dropped the multi-head index $h$ by our i.i.d. assumption. 

Our goal is to find the sample complexity and scaling of feature learning effects with context length ($L$) at finite $d$. To this end, we shall consider the following weight-space version of $\tilde{E}$ given by, 
\begin{align}
\tilde{E}_q &= \sum_{h=1}^H d^2 \mathbb{E}_{A_h \sim q_h}\left[Tr[A_h A_h^\top]-1\right]+\frac{1}{\langle y,K,y\rangle}.
\end{align}
where $K$ is computed using the $q_h$ distribution for $A$. There are several ways of obtaining this variational energy; a direct microscopic route is shown in Sec. \ref{appsec:SoftMaxEDirect}. Alternatively, one can use the arguments of Sec. \ref{App:WeightSpace}, previously employed in Sec. \ref{Sec:CNN}, and trade the kernel version of $\Delta_{l,i}$ with the excess weight decay of that layer. The latter being the first term in the above formula. 

We will focus on two learning patterns here: lazy learning (GP) and weight specialization. For lazy learning, we take $q_h = {\cal N}[0,I_{d^2 \times d^2}/d^2;A_h]$ leading, as expected, to zero contribution from the first term in $\tilde{E}_q$. 

To assess the scaling of $\langle y(X),K(X,X'),y(X')\rangle$ with $L$, we make the following observations. For large $L$ and random $A_h$'s, the denominator of $@_{ab}(X)$ is $O(L)$ plus smaller fluctuations namely 
\begin{align}
@_{ab;h}(X)&= \frac{ e^{[x^a]^\top A_h x^b}}{\sum_{c=1}^L e^{[x^a]^\top A_h x^c}} = \frac{ e^{[x^a]^\top A_h x^b}}{Z+[\sum_{c=1}^L e^{[x^a]^\top A_h x^c}-Z]} \\ \nonumber 
Z &= \mathbb{E}_A \left[\sum_{c=1}^L e^{[x^a]^\top A_h x^c} \right] = O(L) 
\end{align}
which allows an expansion in the small parameter $[\sum_{c=1}^L e^{[x^a]^\top A_h x^c}-Z]/Z=O(1/\sqrt{L})$. Next, we observe that functions involving $l$-sequence index functions, such as our $l=2$ target, appear at $l-2$ order of this expansion. Thus, we suffice with a leading order expansion 
\begin{align}
\langle f(X),y(X) \rangle &= \frac{1}{\sqrt{L}} \sum_{h=1}^H \left\langle Z^{-1} \sum_{ab=1}^L e^{[x^a]^\top A_h x^b} w_h \cdot x^b+O(L^{-1}),\sum_{c,d=1}^L \frac{1}{\sqrt{L (L-1)}}x^c_1 x^d_2 x^d_3 \right \rangle \\ \nonumber 
&=\frac{1}{\sqrt{L}} \sum_{h=1}^H  \left\langle Z^{-1} \sum_{ab=1}^L e^{[x^a]^\top A_h x^b} w_h \cdot x^b+O(L^{-1}), \frac{1}{\sqrt{L (L-1)}}x^a_1 x^b_2 x^b_3 \right \rangle = ... 
\end{align}
where we used the fact that, under our Gaussian measure, $x^d_2 x^d_3$ projects out any term which does not contain an odd power of $x^b_2$ and an odd power of $x^b_3$ and a similar consideration with $x^a_1$. Next, a direct computation gives 
\begin{align}
... &= \frac{1}{\sqrt{L^2 (L-1)}Z} \sum_{h=1}^H Z^{-1} \sum_{ab=1}^L \left([A_h]_{12} [w_h]_3+[A_h]_{13}[w_h]_2(1+O(Tr[A_h A_h^\top])) + O([A_h A_h^\top A_h]_{12} [w_h]_3) + O(A^5)\right) \\ \nonumber &+ O(L^{-1}) = ... 
\end{align}
Noting that under our distribution for $A$, $Tr[A A^\top] = O(1)$ and $O([A_h A^T_h A_h]_{ij})=O([A_h]_{ij})$ we finally obtain that the typical scale of the r.h.s. the same as its leading order expansion in $A$ namely  
\begin{align}
\label{AppEq:AlignmentScale}
... &\propto  \frac{\sqrt{L}}{Z}\sum_{h=1}^H \left([A_h]_{12} [w_h]_3+[A_h]_{13}[w_h]_2\right)
\end{align}
Finally, computing $\langle y(X),K(X,X'),y(X')\rangle$ amount to computing the 2nd moment of the above term under the prior for $A_h,w_h$ which, given that $Z = O(L)$, yields a scaling  
\begin{align}
\label{Eq:TransformerGPKernel}
\langle y,K,y\rangle &\propto \frac{1}{L d^3}
\end{align}
Implying that $\tilde{E}_{lazy} \propto L d^3$.

We next consider a specializing scenario ($q_{sp.}$) where for $O(1)$ heads, $[A_h]_{12}$ has an average of the order $M/d$. Other components of $A_h$ as well as $A_h$'s of the remaining heads, remain lazy. The motivation for this choice, also observed empirically, is that together with a spike in $[w_h]_3$, one can create strong alignment in Eq. (\ref{AppEq:AlignmentScale}). Other combinations of $1,2,3$ indices are also possible and lead to the same scaling. 

Using the above $q_{sp.}$, the excess weight-decay (first term in $\tilde{E}_q$) gets contributions only from the $O(1)$ specializing heads scaling as $d^2 (M/d)^2$. One further notes that using this $A_{12}$, in Eq. \ref{AppEq:AlignmentScale} and subsequently in \ref{Eq:TransformerGPKernel}, yields $\langle y,K,y\rangle \propto \frac{M^2/d^2}{Ld H}$ leading to $\tilde{E}_{sp.}= M^2 + H Ld^3/M^2$. Optimizing the latter over $M^2$ leads to $\tilde{E}_{sp,opt}=\sqrt{L H} d^{1.5}$. This scenario is thus favorable to lazy learning up to $H=O(\sqrt{L d^3})$.  

To validate these results, we consider the above model, with $ d=8$ and $d=16$, $ H=2$, $ \kappa=0.1$, and $ P=350, 500, 707, 1000$ data points, and a maximal context length of $ 180, 360, 720, 720$ respectively. Fig. \ref{fig:Alignment_specialization}(b) shows target alignment of the equilibrated network as a function of $\sqrt{Ld^3}/P$, demonstrating that $P=\sqrt{Ld^3}$ is the scale at which $O(1)$ alignment appears. A similar plot for the Test MSE results is shown in the right panel. 

\subsubsection{Direct derivation of $\tilde{E}_q$ for a softmax block}
\label{appsec:SoftMaxEDirect}
Gathering all $A_h,w$ parameters into $\Theta$, the probability of aligment $\alpha$ is given by 
\begin{equation}
    p_{A_f}(\alpha) = \frac{1}{2\pi}\int dt\int d\Theta p\left(\Theta\right)\exp\left(it\left(\left\langle y,f_\Theta \right\rangle -\alpha\right)\right).
\end{equation}
We can integrate out $w_h$ as follows:
\begin{align*}
    p_{A_f}(\alpha) &= \frac{1}{2\pi}\int dt\int d\Theta p\left(\Theta\right)\exp\left(it\left(\left\langle y,\frac{1}{\sqrt{L}} \sum_{h=1}^H \sum_{a,b=1}^L @_{ab;h}(x) (w_h \cdot x^b) \right\rangle -\alpha\right)\right).
\end{align*}
Let $v_h(x) := \sum_{b=1}^L \left(\sum_{a=1}^L @_{ab;h}(x)\right) x^b$. Then, we can re-write
\begin{align*}
    p_{A_f}(\alpha) &= \frac{1}{2\pi}\int dt\int d\Theta p\left(\Theta\right)\exp\left(it\left(\frac{1}{\sqrt{L}} \sum_{h=1}^H \left\langle y, w_h \cdot v_h \right\rangle -\alpha\right)\right) \\
    &= \frac{1}{2\pi}\int dt\int d\Theta p\left(\Theta\right)\exp\left(it\left(\frac{1}{\sqrt{L}} \sum_{h=1}^H w_h \cdot \left\langle y, v_h \right\rangle -\alpha\right)\right)
\end{align*}
where we're using an unusual notation, regarding $\left\langle y, v_h \right\rangle$ as a vector of $L_2$ inner products. Using the prior we continue:
\begin{align*}
    &= \frac{1}{2\pi}\int dt \exp(-it\alpha) \prod_{h=1}^H \int dA_h  p\left(A_h\right) \int dw_h  p\left(w_h\right) \exp\left(\frac{it}{\sqrt{L}} w_h \cdot \left\langle y, v_h \right\rangle \right) \\
    &= \frac{1}{2\pi}\int dt \exp(-it\alpha) \prod_{h=1}^H \int dA_h  p\left(A_h\right) \exp\left(-\frac{1}{2}\frac{t^2}{LdH} \left\langle y, v_h \right\rangle^\top \left\langle y, v_h \right\rangle\right) \\
    &= \frac{1}{2\pi}\int dA\, p(A) \int dt  \exp\left(-\frac{1}{2}\frac{t^2}{LdH} \sum_{h=1}^H \left\langle y, v_h \right\rangle^\top \left\langle y, v_h \right\rangle -it\alpha \right)
\end{align*}
When we integrate out $t$ we get
\begin{align}
& p_{A_f}(\alpha) = \\
& = \frac{1}{\sqrt{2\pi}} \int dA\, p(A) \exp\left(-\frac{\alpha^{2}}{\frac{2}{LdH}\sum_{h=1}^{H} \left\langle y, v_h \right\rangle^\top \left\langle y, v_h \right\rangle}+\frac{1}{2}\log\frac{1}{\frac{1}{LdH}\sum_{h=1}^{H} \left\langle y, v_h \right\rangle^\top \left\langle y, v_h \right\rangle}\right)\nonumber 
\end{align}
Let's denote
\begin{align*}
    \kappa(A) := \frac{1}{LdH}\sum_{h=1}^{H} \left\langle y, v_h \right\rangle^\top \left\langle y, v_h \right\rangle = \frac{1}{LdH}\sum_{h=1}^{H} \sum_{i=1}^d \left\langle y, \left[v_h\right]_i \right\rangle^2.
\end{align*}
Then we have
\begin{align*}
    p_{A_f}(\alpha) &= \int dA\, p(A) \frac{1}{\sqrt{2\pi \kappa(A)}} \exp\left(-\frac{\alpha^{2}}{2\kappa(A)}\right) \\
    &= \int dA\, p(A) \gN(0,\kappa(A);\alpha)
\end{align*}

Next, we take a layer-decoupling approximation, similar in spirit to kernel adaptation, where we assume $\kappa(A)$ does not fluctuate out of scale, with $A$. We thus trade $\kappa(A)$ by its average, which yields our $a_y$ term, $\alpha^2/\langle y, K , y \rangle$. As in the original derivation, we expect $a_y$ to be extensive ($O(d,N,..)$) and hence its log, or equivalently the above $1/\sqrt{2\pi \kappa(A)}$ prefactor of the exponent is negligible. The action ($S$), or negative log-probability associated with $p(A)e^{-\alpha^2/(2\kappa(A))}$ is now given by $S=(d^2 Tr[A^T A]+\alpha^2/\kappa(A))/2$. Using a variational principle on this action with a Gaussian $q$, one obtains the above variational energy, with the $-d^2$ contribution coming from the entropy term $\int dA q(A) \log(q(A))$ in the KL-divergence used in the variational approach (similarly to how $-H_{q,\alpha}(h)$ emerged in sec. \ref{app:variation_est_intro})

\section{Experimental details} \label{App:2layer}

\subsubsection{Two-layer FCN- Experimental Details}

In panels (a) and (b) of Fig.~\ref{fig:2_layer_exp}, the experimental data points were computed by training an ensemble of 300 networks. For panel (a), all experiments used $P=40d$ and $N=d=40$, with a ridge parameter of $1$ and variances of $\sigma_1^2=\sigma_2^2=1$. For panel (b), we set $N=d$ while varying both $d$ and $P$. The experimental value of $P_*$ was calculated by extrapolating from the different values of $P$ and their corresponding alignments, and then inverting this correspondence to find the value of $P$ for which $A_f=0.1$.

In panel (c), we set $P=40d$ and $d=40$, while increasing the value of $N$. As in the other panels, we used $\sigma_1^2=\sigma_2^2=1$, but here we set the ridge parameter to $\kappa=0.25$. We trained an ensemble of 300 networks and computed a histogram of the pre-activation alignment in the hidden layer. First-layer outliers were defined as any pre-activations having an overlap with $\Phi_*(x) = w_* \cdot x$ of more than $3\sigma_1/\sqrt{d}$ (three times the standard deviation of the GP distribution). The total number of such outliers was counted and then averaged over the ensemble of networks.

\subsection{Three-layer FCN- Experimental Details}
\label{app:3layer}
\begin{figure}
    \includegraphics[width=1.1\linewidth]{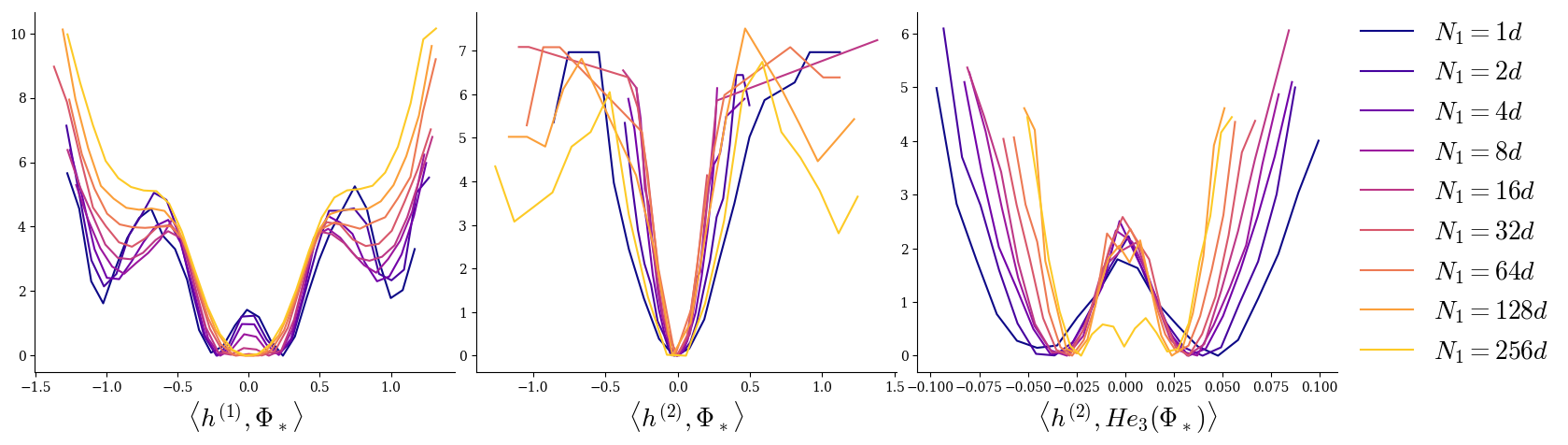}
    \caption{Distribution of pre-activations of the alignment of the first and second layers with the linear feature- $\Phi_*=w_*\cdot x$, and the cubic feature- $\hat{H}e_3(\Phi_*)$ where the hat simply denotes the normalized third Hermite polynomial. As can be seen in this figure, increasing the width of the first layer-$N_1$ pushes the preferred feature learning pattern, from Specialization-magnetization to GP-Specialization, as predicted by the our approach. }
    \label{fig:FCN3_phase_change}
\end{figure}

In panel (a) of Fig.~\ref{fig:Alignment_specialization}, the experimental data was computed by training an ensemble of 30 networks. For all experiments, we set $d=N_1=N_2$, with a ridge parameter of $0.125$ and variances of $\sigma_1^2=\sigma_2^2=0.25$.

For panel (c) of Fig.~\ref{fig:Alignment_specialization} and in Fig.~\ref{fig:FCN3_phase_change}, we set $P=40d$ and $d=N_2=10$, while increasing the value of $N_1$. We further used $\sigma_1^2=\sigma_2^2=0.25$ and a ridge parameter of $\kappa=0.125$. An ensemble of 300 networks was trained to compute the following histograms for the pre-activation alignments: 1. of the linear feature in the first hidden layer, 2. with the linear feature in the second layer, and 3. with the target in the second layer. Outliers in the first layer were calculated as in the two-layer case. For the second layer, activation alignments deviating by more than 3 times the NNGP standard deviation were considered outliers.

\subsection{Alignment as a tight bound on MSE}

A central component of our approach is using alignment as a certificate for good learning. As discussed in the main body of the text, the alignment is indeed a lower bound on the MSE through the inequality MSE$\geq (1-A_f)^2$. However, this inequality does not necessitate that the MSE will vanish. We therefore provide in this section empirical evidence that strong alignment is indeed an indication of MSE. To this extent, we provide results both from the softmax attention experiments and the three layer case, and in both we find that strong alignment is closely correlated with low loss, as can be seen in Fig. 
\begin{figure}
    \centering
    \includegraphics[width=1\linewidth]{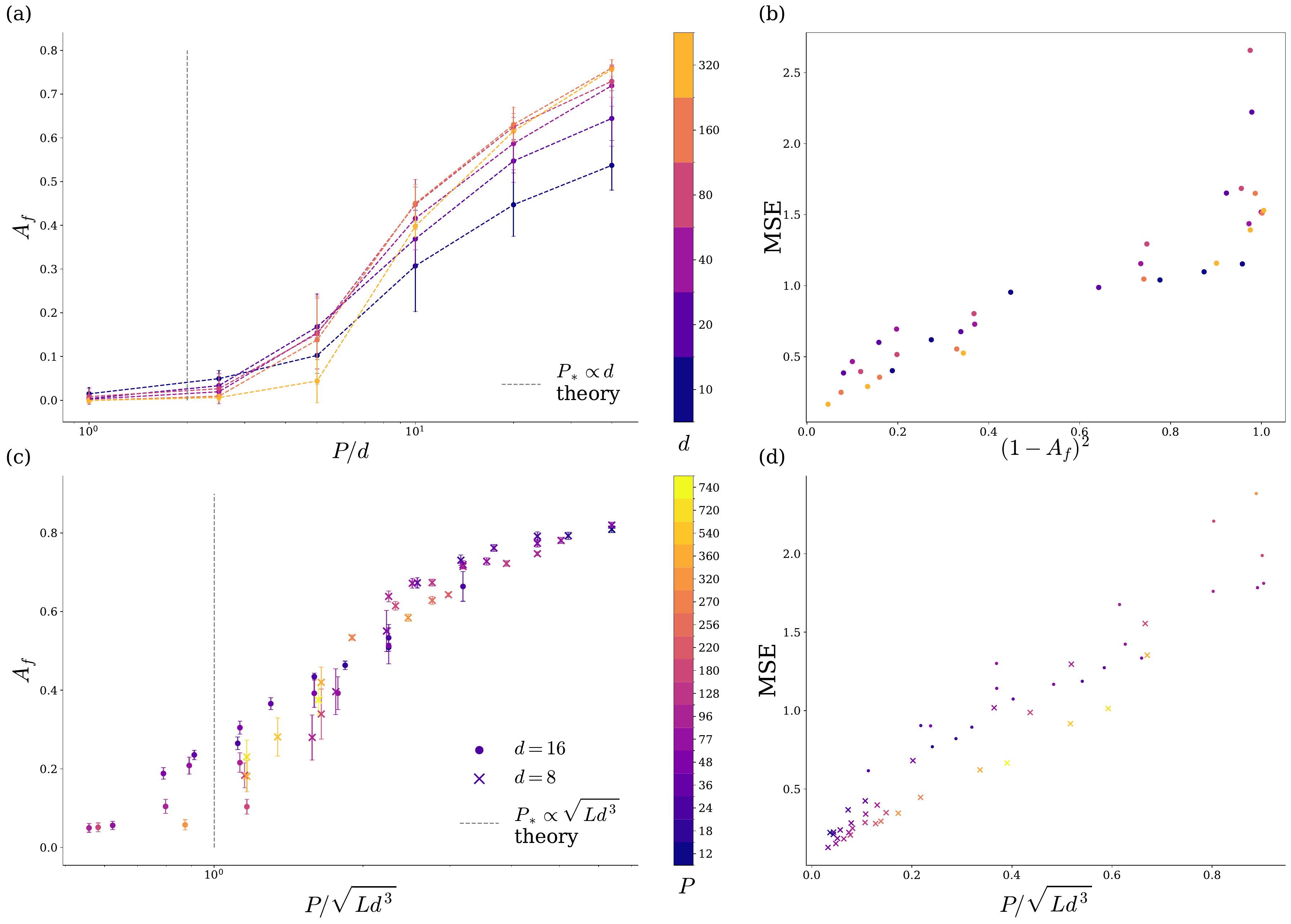}
    \caption{Evidence that alignment is a tight lower bound on MSE. Here we provide results from two experiments, softmax attention (panels(c) and (d)), and three layer network (panels (a) and (b)). These results clearly imply that the alignment is a strong indicator of good MSE. }
    \label{appfig:mse}
\end{figure}

\section{Comparing Kernel Adaptation and LDT}\label{app:posterior_mean}

The results in the heuristic approach presented here are closely related to the kernel adaptation approach \cite{fischer_critical_2024,seroussi_separation_2023,ringel_applications_2025,rubin_grokking_2024,rubin_kernels_2025}. In this section, we compute explicitly the mean posterior predictor for the alignment. Interestingly, we find that the equations for the LDT approach are remarkably similar to those of the mean predictor. 

As in the derivation of the Chernhof bounds we begin by integrating out the readout weights from the posterior, and introducing a delta function Fourier field in the prior, we obtain-
\begin{equation}
Z=\int d\tilde{t}\int d\Theta^{L-1}p\left(\Theta^{L-1}\right)\exp\left(-\frac{\sigma_{L}^{2}}{2N_{L}\chi}\sum_{i=1}^{N_{L}}\left\langle \sigma\left(h_{i}^{L-1}\left(x\right)\right),\tilde{t}\left(x\right)\right\rangle ^{2}-\frac{\kappa}{2}\left\langle \tilde{t},\tilde{t}\right\rangle +i\left\langle y,\tilde{t}\right\rangle \right)
\end{equation}
where $\tilde{t}$ is an imaginary helper field originating from the
delta function. Define- 
\begin{equation}
\tilde{t}\left(x\right)=t_{y}y\left(x\right)+t_{\perp}y_{\perp}\left(x\right)
\end{equation}

where $y_{\perp}$ is some function that is orthogonal to $y$, so
that $\left\langle y,y_{\perp}\right\rangle =0$. Now we have- 
\begin{align}
Z & =\int dt_{y}dt_{\perp}\exp\left(-\frac{\kappa}{2}\left(t_{y}^{2}+t_{\perp}^{2}\right)+it_{y}\right)\int d\Theta^{L-1}p\left(\Theta^{L-1}\right)\\
 & \ \ \ \ \cdot\exp\left(-\sum_{i=1}^{N_{L}}\frac{\sigma_{L}^{2}}{2N_{L}}\left(t_{y}\left\langle y\left(x\right),\sigma\left(h_{i}^{L-1}\left(x\right)\right)\right\rangle +t_{\perp}\left\langle y_{\perp}\left(x\right),\phi\left(w_{i}\cdot x\right)\right\rangle \right)^{2}\right)\nonumber 
\end{align}
Next, we assume self consistantly that $t_{\perp}$ vanishes, and
so the above equation simplifies to: 
\begin{align}
Z & =\int dt_{y}dt_{\perp}\exp\left(-\frac{\kappa}{2}\left(t_{y}-\frac{i}{\kappa}\right)^{2}\right)\int d\Theta^{L-1}p\left(\Theta^{L-1}\right)\exp\left(-\sum_{i=1}^{N_{L}}\frac{\sigma_{L}^{2}}{2N_{L}}t_{y}^{2}\underbrace{\left\langle y\left(x\right),\sigma\left(h_{i}^{L-1}\left(x\right)\right)\right\rangle ^{2}}_{a_{i}\left(\Theta^{L-1}\right)}\right)
\end{align}
Here enters the adaptive approach. We assume that $t_{y}$ is weakly
fluctuating, and approximate- 
\begin{equation}
\exp\left(\frac{\sigma_{a}^{2}}{2N}t_{y}^{2}A_{i}^{2}\right)\approx\exp\left(-\sum_{i=1}^{N_{L}}\frac{\sigma_{L}^{2}}{2N_{L}}\mathbb{E}_{t_{y}}\left[t_{y}^{2}\right]a_{i}^{2}\left(\Theta^{L-1}\right)+\sum_{i=1}^{N_{L}}\frac{\sigma_{L}^{2}}{2N_{L}}t_{y}^{2}\mathbb{E}_{\Theta^{L-1}}\left[a_{i}^{2}\right]\right)
\end{equation}
So that 
\begin{align}
Z & \propto\int dt_{y}\exp\left(\underbrace{-\frac{\kappa}{2}\left(t_{y}-\frac{i}{\kappa}\right)^{2}-\sum_{i=1}^{N_{L}}\frac{\sigma_{L}^{2}}{2N_{L}}t_{y}^{2}\mathbb{E}_{\Theta^{L-1}}\left[a_{i}^{2}\right]}_{:=-S_{t_{y}}}\right)\\
 & \cdot\int d\Theta^{L-1}\exp\left(\underbrace{-\sum_{i=1}^{N_{L}}\frac{\sigma_{L}^{2}}{2N_{L}}\mathbb{E}_{t_{y}}\left[t_{y}^{2}\right]a_{i}^{2}\left(\Theta^{L-1}\right)+\log p\left(\Theta^{L-1}\right)}_{:=-S_{\Theta^{L-1}}}\right)\nonumber 
\end{align}
We require self consistently that $\mathbb{E}_{\Theta^{L-1}}\left[a_{i}\left(\Theta^{L-1}\right)\right]$
is indeed given by the expectation with respect to the action $S_{\Theta^{L-1}}$
\begin{align}
\mathbb{E}_{\Theta^{L-1}}\left[a_{j}\left(\Theta^{L-1}\right)\right]=\mathbb{E}_{\Theta^{L-1}\sim S_{\Theta^{L-1}}}\left[a_{j}\left(\Theta^{L-1}\right)\right]\\
=\frac{\int d\Theta^{L-1}p\left(\Theta^{L-1}\right)a_{j}\left(\Theta^{L-1}\right)\exp\left(-\frac{\sigma_{L}^{2}}{2N_{L}}\mathbb{E}_{t_{y}}\left[t_{y}^{2}\right]\sum_{i=1}^{N_{L}}a_{i}\left(\Theta^{L-1}\right)\right)}{\int d\Theta^{L-1}p\left(\Theta^{L-1}\right)\exp\left(-\sum_{i=1}^{N_{L}}\frac{\sigma_{L}^{2}}{2N_{L}}\mathbb{E}_{t_{y}}\left[t_{y}^{2}\right]\sum_{i=1}^{N_{L}}a_{i}\left(\Theta^{L-1}\right)\right)}\nonumber 
\end{align}

Denoting- $\sum_{i=1}^{N_{L-1}}a_{i}\left(\Theta^{L-1}\right):=a\left(\Theta^{L-1}\right)$,
and $\int d\Theta^{L-1}p\left(\Theta^{L-1}\right)\left(...\right)=\mathbb{E}_{\Theta^{L-1}\sim\text{GP}}\left[\left(...\right)\right]$,
then we have- 
\begin{align}
\sum_{j=1}^{N_{L}}\mathbb{E}_{\Theta^{L-1}\sim S_{\Theta^{L-1}}}\left[a_{j}\left(\Theta^{L-1}\right)\right] & :=\mathbb{E}_{\Theta^{L-1}\sim S_{\Theta^{L-1}}}\left[a\left(\Theta^{L-1}\right)\right]\\
 & =\frac{\mathbb{E}_{\Theta^{L-1}\sim\text{GP}}\left[a\left(\Theta^{L-1}\right)\exp\left(-\frac{\sigma_{L}^{2}}{2N_{L}}\mathbb{E}_{t_{y}}\left[t_{y}^{2}\right]a\left(\Theta^{L-1}\right)\right)\right]}{\mathbb{E}_{\Theta^{L-1}\sim\text{GP}}\left[\exp\left(-\frac{\sigma_{a}^{2}}{2N}\overline{t}_{y}^{2}a\left(\Theta^{L-1}\right)\right)\right]}\nonumber 
\end{align}

Next we require that the same condition holds for $t_{y}$. Note that
the action for $t_{y}$ is simply a Gaussian one, and following square
completion we have-
\begin{equation}
e^{-S_{t_{y}}}\propto\exp\left(-\frac{\left(\kappa+\frac{\sigma_{L}^{2}}{N_{L}}\mathbb{E}_{\Theta^{L-1}}\left[a\left(\Theta^{L-1}\right)\right]\right)}{2}\left(t_{y}-\frac{i}{\left(\kappa+\frac{\sigma_{L}^{2}}{N_{L}}\mathbb{E}_{\Theta^{L-1}}\left[a\left(\Theta^{L-1}\right)\right]\right)}\right)^{2}\right)
\end{equation}
So that 
\begin{equation}
t_{y}\sim\mathcal{N}\left(i\left(\kappa+\frac{\sigma_{L}^{2}}{N_{L}}\mathbb{E}_{\Theta^{L-1}}\left[a\left(\Theta^{L-1}\right)\right]\right)^{-1},\left(\kappa+\frac{\sigma_{L}^{2}}{N_{L}}\mathbb{E}_{\Theta^{L-1}}\left[a\left(\Theta^{L-1}\right)\right]\right)^{-1}\right)
\end{equation}
So that we can substitute the self consistency requirement on the
$\Theta$ variables into the expression for the mean of $t_{y}$ and
we obtain
\[
\mathbb{E}_{t_{y}\sim S_{t_{y}}}\left[t_{y}\right]=i\left(\kappa+\frac{\frac{\sigma_{L}^{2}}{N_{L}}\mathbb{E}_{\Theta^{L-1}\sim\text{GP}}\left[a\left(\Theta^{L-1}\right)\exp\left(-\frac{\sigma_{L}^{2}}{2N_{L}}\mathbb{E}_{t_{y}\sim S_{t_{y}}}\left[t_{y}^{2}\right]a\left(\Theta^{L-1}\right)\right)\right]}{\mathbb{E}_{\Theta^{L-1}\sim\text{GP}}\left[\exp\left(-\frac{\sigma_{a}^{2}}{2N}\mathbb{E}_{t_{y}\sim S_{t_{y}}}a\left(\Theta^{L-1}\right)\right)\right]}\right)^{-1}
\]
Note that if we approximate- $\mathbb{E}_{t_{y}\sim S_{t_{y}}}\left[t_{y}^{2}\right]\sim\mathbb{E}_{t_{y}\sim S_{t_{y}}}^{2}\left[t_{y}\right]$
, then we obtain a self consistent equation for $\mathbb{E}_{t_{y}\sim S_{t_{y}}}\left[t_{y}\right]:=\overline{t}_{y}$,
given by-
\begin{equation}
\overline{t}_{y}=\frac{i\mathbb{E}_{\Theta^{L-1}\sim\text{GP}}\left[\exp\left(-\frac{\sigma_{L}^{2}}{2N_{L}}\overline{t}_{y}^{2}a\left(\Theta^{L-1}\right)\right)\right]}{\mathbb{E}_{\Theta^{L-1}\sim\text{GP}}\left[\left(\kappa+\frac{\sigma_{L}^{2}}{N_{L}}a\left(\Theta^{L-1}\right)\right)\exp\left(-\frac{\sigma_{L}^{2}}{2N_{L}}\overline{t}_{y}^{2}a\left(\Theta^{L-1}\right)\right)\right]}
\end{equation}
Rewriting with real $\overline{t}_{y}\mapsto i\overline{t}_{y}$,
we obtain the self consistency equation- 
\begin{equation}
\overline{t}_{y}=\frac{\mathbb{E}_{\Theta^{L-1}\sim\text{GP}}\left[\exp\left(\frac{\sigma_{L}^{2}}{2N_{L}}\overline{t}_{y}^{2}a\left(\Theta^{L-1}\right)\right)\right]}{\mathbb{E}_{\Theta^{L-1}\sim\text{GP}}\left[\left(\kappa+\frac{\sigma_{L}^{2}}{N_{L}}a\left(\Theta^{L-1}\right)\right)\exp\left(\frac{\sigma_{L}^{2}}{2N_{L}}\overline{t}_{y}^{2}a\left(\Theta^{L-1}\right)\right)\right]}
\end{equation}

Replacing $\tilde{a}=\kappa+\frac{\sigma_{L}^{2}}{N_{L}}a\left(\Theta^{L-1}\right)$,
we obtain the following self consistent equation for $\overline{t}_{y}$:
\begin{align}
\overline{t}_{y} & =\frac{\mathbb{E}_{\Theta^{L-1}\sim\text{GP}}\left[\exp\left(\frac{\sigma_{L}^{2}}{2N_{L}}\overline{t}_{y}^{2}\tilde{a}\left(\Theta^{L-1}\right)\right)\right]}{\mathbb{E}_{\Theta^{L-1}\sim\text{GP}}\left[\left(\frac{\sigma_{L}^{2}}{N_{L}}\tilde{a}\left(\Theta^{L-1}\right)\right)\exp\left(\frac{\sigma_{L}^{2}}{2N_{L}}\overline{t}_{y}^{2}\tilde{a}\left(\Theta^{L-1}\right)\right)\right]}
\end{align}
This expression is very similar to the one we found for the upper Chernoff bound, and the two equations coincide in the limit $\kappa\rightarrow0$, and $\alpha\rightarrow1$.


\end{document}